%% file: main.tex
\documentclass[letterpaper, 10 pt, journal, twoside]{IEEEtran}
\usepackage{times}
\newcommand{\citep}{\cite}
\newcommand{\citet}{\cite}
\usepackage{multicol}
\usepackage{xcolor}
\usepackage[bookmarks=true]{hyperref}
\hypersetup{letterpaper,bookmarksopen,bookmarksnumbered,
breaklinks=true,
pdfpagemode=UseOutlines,
colorlinks=true,
linkcolor=blue,
anchorcolor=blue,
citecolor=red,
filecolor=blue,
menucolor=blue,
urlcolor=blue
}
\usepackage{amssymb,amsmath}
\usepackage{gensymb}
\usepackage{cite} 
\usepackage{balance}
\usepackage{booktabs}
\usepackage{caption}
\usepackage[font=footnotesize,labelformat=simple]{subcaption}
\usepackage[font=footnotesize, skip=0pt]{caption}
\usepackage{graphicx}
\usepackage{tabularx}
\usepackage{pifont}

\usepackage{dsfont}
\usepackage{svg}

\newcommand{\figref}[1]{Fig.\ref{#1}}
\usepackage{amsthm} 
\usepackage{algorithm}
\usepackage{algorithmicx}
\usepackage[noend]{algpseudocode}
\usepackage{amsmath}
\usepackage{svg}
\usepackage{xspace}
\usepackage[normalem]{ulem}
\usepackage{multirow}
\usepackage{tabularx}
\usepackage{subcaption}
\let\labelindent\relax
\usepackage{enumitem}
\usepackage{tikz}
\usetikzlibrary{fit}

\usepackage{mdframed}

\newcommand\algname[1]{\textsf{#1}\xspace}
\newcommand\astar{\algname{A*}}
\newcommand\mrastar{\algname{MRA*}}

\newcommand\hastar{\algname{HA*}}
\newcommand\ighastar{\algname{IGHA*}}
\newcommand\ihastar{\algname{HA*M}}
\newcommand\RRTstar{\algname{RRT*}}
\newcommand\RRT{\algname{RRT}}
\newcommand\drk{\ighastar-$\infty$~}
\newcommand\dsrk{\ighastar-$0$~}

\newcommand\activate{\textsc{Activate}\xspace}
\newcommand\shift{\textsc{Shift}\xspace}

\newcommand{\xxnote}[3]{}
\ifx\hidenotes\undefined
  \renewcommand{\xxnote}[3]{\color{#2}{#1: #3}}
\fi
\algdef{SE}[DOWHILE]{Do}{DoWhile}{\algorithmicdo}[1]{\algorithmicwhile\ #1}%

\newtheorem{definition}{Definition}[section]
\newtheorem{theorem}{Theorem}[section]

\newtheorem{corollary}{Corollary}[section]

\newtheorem{observation}{Observation}[section]
\DeclareMathOperator{\Succ}{\textsc{Succ}}

\begin{document}

\title{Incremental Generalized Hybrid A*}
\author{
    Sidharth Talia$^{1}$,
    Oren Salzman$^{2}$, and
    Siddhartha Srinivasa$^{1}$
\thanks{Manuscript received: August, 7, 2025; Revised October, 20, 2025; Accepted November, 22, 2025.
This paper was recommended for publication by Editor Olivier Stasse upon evaluation of the Associate Editor and Reviewers' comments.
This work was (partially) funded by the National Science Foundation 
NRI (\#2132848) and CHS (\#2007011), DARPA RACER (\#HR0011-21-C-0171), the Office of Naval Research (\#N00014-17-1-2617-P00004 and \#2022-016-01 UW), Amazon, Technion Autonomous Systems Program (TASP), and the United States-Israel Binational Science Foundation (BSF) grant (\#2021643).
$^{1}$University of Washington, Seattle, USA.~{\tt\footnotesize~\{sidtalia, siddh\}@cs.washington.edu}, 
$^{2}$Technion-Israel Institute of Technology, Haifa, Israel. {\tt\footnotesize osalzman@cs.technion.ac.il}.
Digital Object Identifier (DOI): see top of this page.}
}
\markboth{IEEE Robotics and Automation Letters. Preprint Version. Accepted Nov, 2025}
{Talia \MakeLowercase{\textit{et al.}}: Incremental Generalized Hybrid \astar} 

\maketitle

\begin{abstract}
We address the problem of efficiently organizing search over very large trees, which arises in many applications ranging from autonomous driving to aerial vehicles. 
Here, we are motivated by off-road autonomy, where real-time planning is essential.
Classical approaches use graphs of motion primitives and exploit dominance to mitigate the curse of dimensionality and prune expansions efficiently.
However, for complex dynamics, repeatedly solving two-point boundary-value problems makes graph construction too slow for fast kinodynamic planning.
Hybrid A* (\hastar) addressed this challenge by searching over a tree of motion primitives and introducing approximate pruning using a grid-based dominance check.
However, choosing the grid resolution is difficult: too coarse risks failure, while too fine leads to excessive expansions and slow planning.
We propose Incremental Generalized Hybrid A* (\ighastar), an anytime tree-search framework that dynamically organizes vertex expansions without rigid pruning. 
\ighastar provably matches or outperforms \hastar.
For both on-road kinematic and off-road kinodynamic planning queries for a car-like robot, variants of \ighastar use $6\times$ fewer expansions to the best solution compared to an optimized version of \hastar (\ihastar, an internal baseline).
In simulated off-road experiments in a high-fidelity simulator, \ighastar outperforms \ihastar when both are used in the loop with a model predictive controller.
We demonstrate real-time performance both in simulation and on a small-scale off-road vehicle, enabling fast, robust planning under complex dynamics. 
Website: \url{https://personalrobotics.github.io/IGHAStar/}
\end{abstract}
\vspace{5pt}

\IEEEpeerreviewmaketitle

\input{sections/Introduction}
\begin{figure*}
\centering
\begin{subfigure}{0.195\linewidth}
  \centering
  \includegraphics[width=\linewidth]{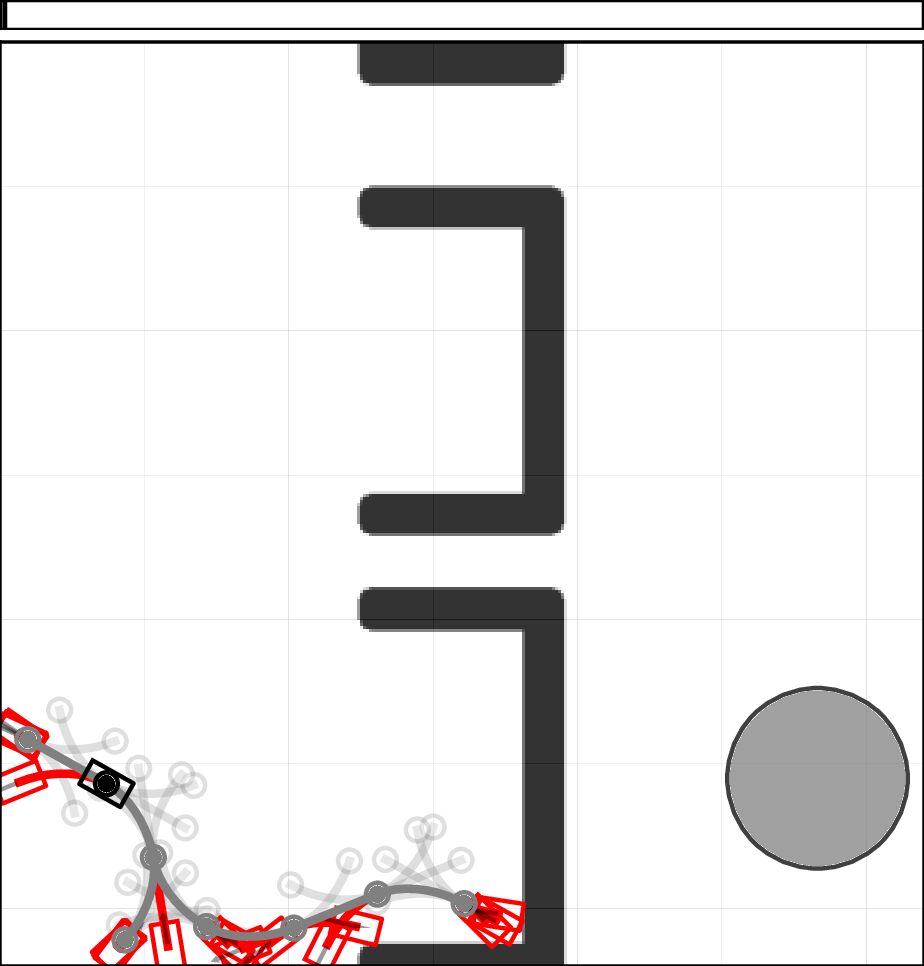}
  \caption{1$\times$Resolution}
  \label{fig: main_0}
\end{subfigure}
\begin{subfigure}{0.195\linewidth}
  \centering
  \includegraphics[width=\linewidth]{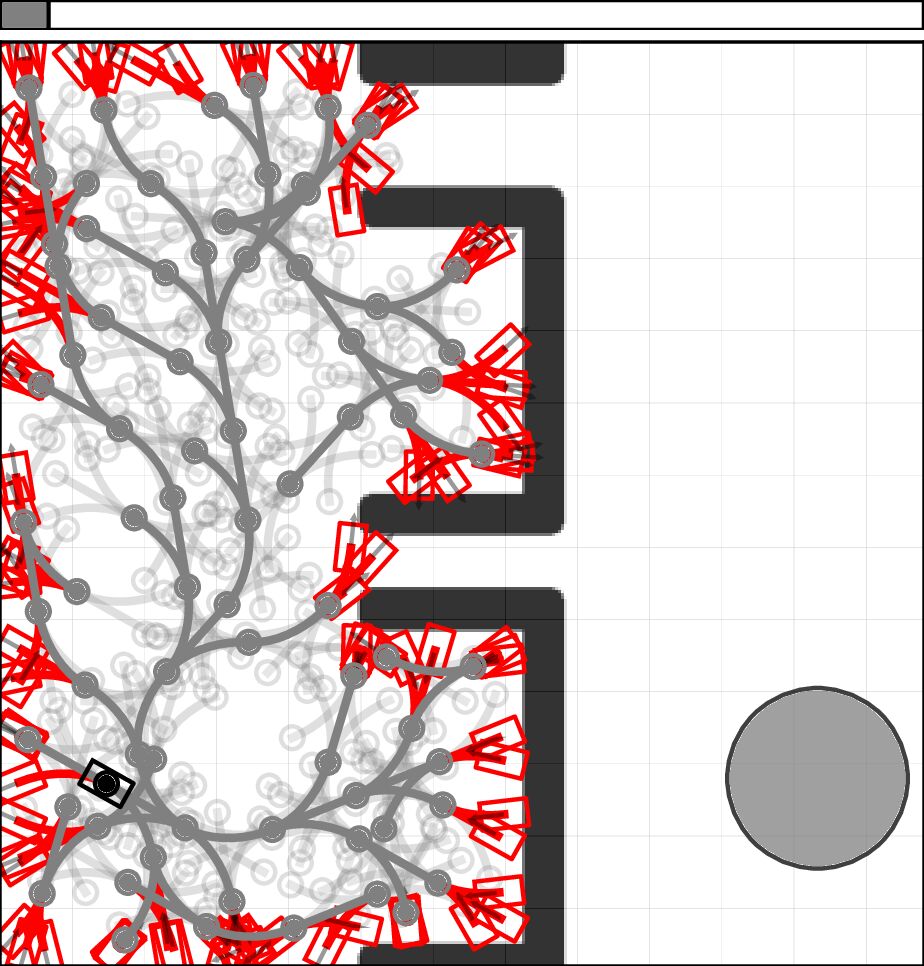}
  \caption{2$\times$Resolution}
  \label{fig: main_1}
\end{subfigure}
\begin{subfigure}{0.195\linewidth}
  \centering
  \includegraphics[width=\linewidth]{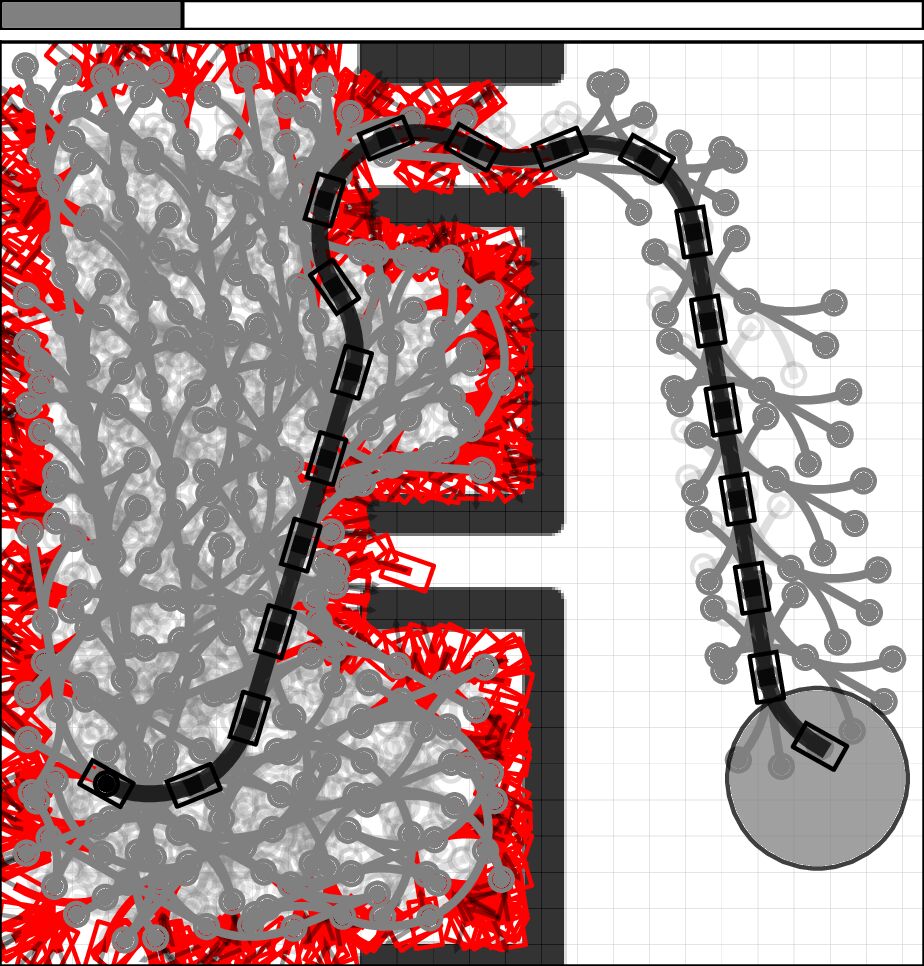}
  \caption{4$\times$Resolution}
  \label{fig: main_2}
\end{subfigure}
\begin{subfigure}{0.195\linewidth}
  \centering
  \includegraphics[width=\linewidth]{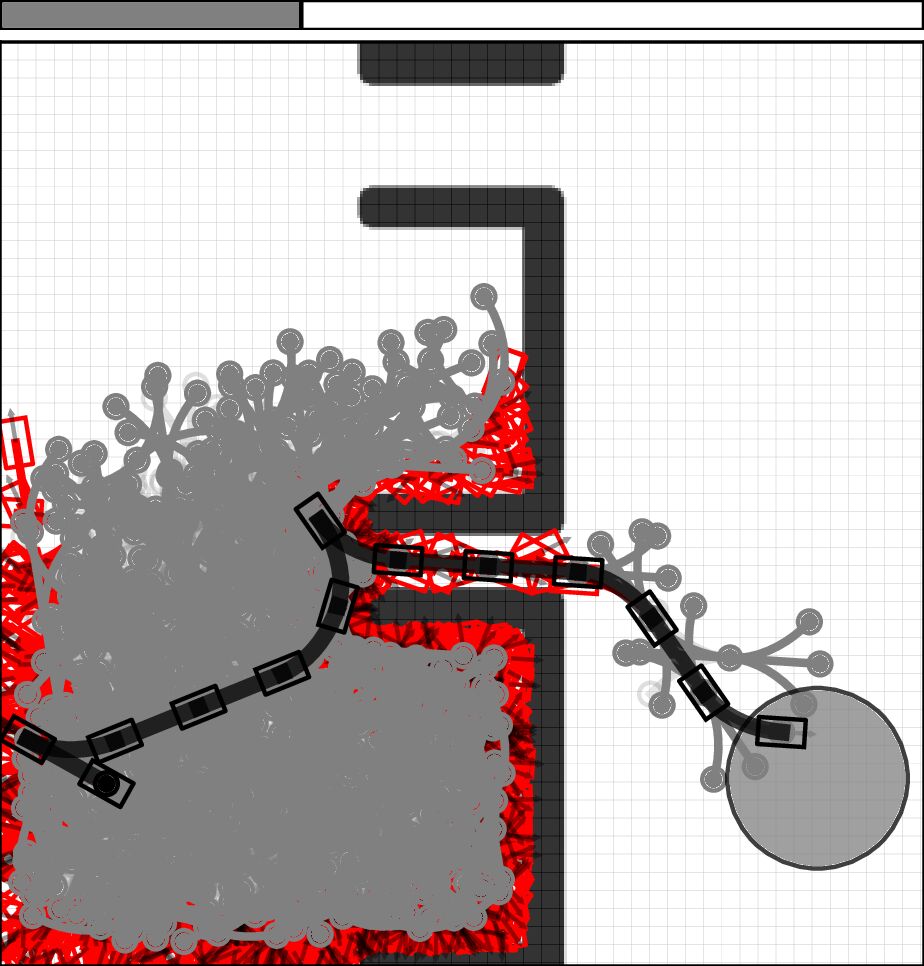}
  \caption{8$\times$Resolution}
  \label{fig: main_3}
\end{subfigure}
\begin{subfigure}{0.195\linewidth}
  \centering
  \includegraphics[width=\linewidth]{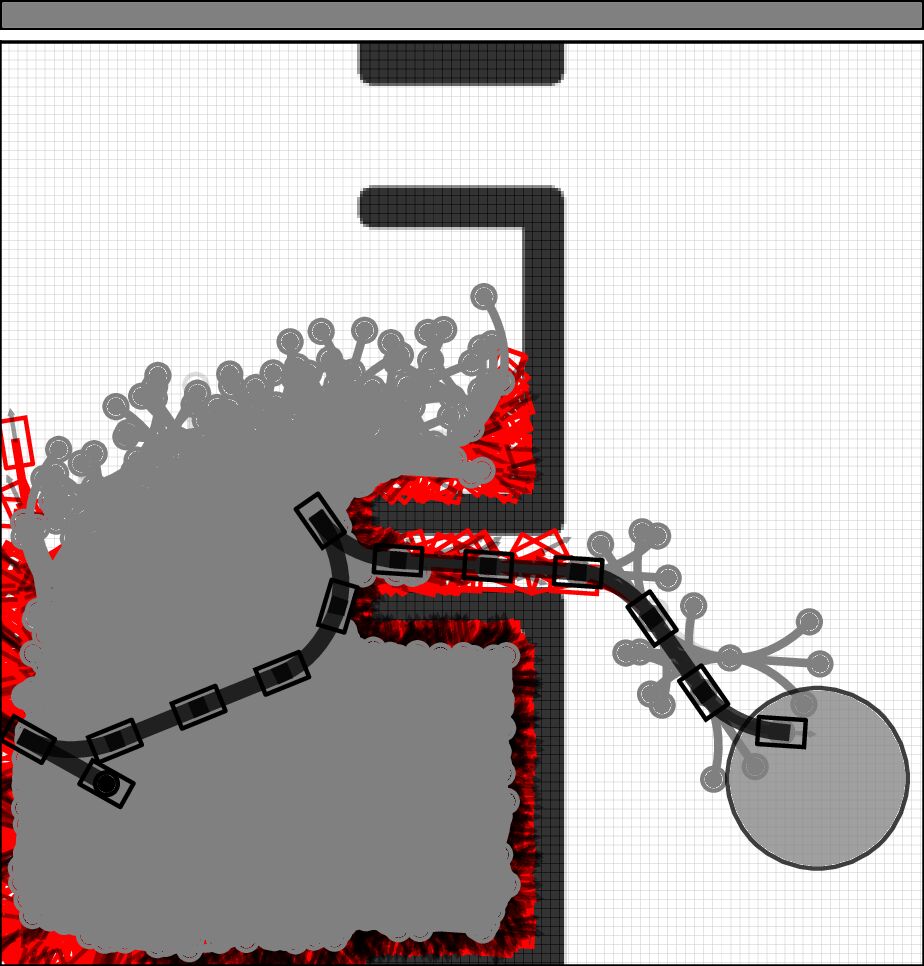}
  \caption{16$\times$Resolution}
  \label{fig: main_4}
\end{subfigure}
\caption{
\hastar operating at different resolutions for the same planning query.
Vertices found (Open list) are shown in gray, pruned vertices in light gray, invalid (collision) in red, the goal set is marked as the circular region on the bottom right, the path is shown in black, and the gray bar on top represents the number of expansions.
When using \hastar, we don't a priori know the resolution for the first or the best solution, as shown in~\subref{fig: main_2},~\subref{fig: main_3}.
As a result, if the discretization is too coarse, we can end up with no solution as shown in~\subref{fig: main_0} and~\subref{fig: main_1} respectively.
If it is too fine, we expand more than necessary as shown in~\subref{fig: main_4}, where the solution quality does not improve compared to~\subref{fig: main_3} despite $\approx3\times$ the number of expansions in the former.
}
\label{fig: main}
\vspace{-15pt}
\end{figure*}

\input{sections/literature}
\input{sections/Problem}
\input{sections/Method}
\input{sections/Experiments}
\input{sections/Future_Work}

\footnotesize
{\bibliographystyle{IEEEtran}
\bibliography{references}
}

\end{document}

%% file: sections/Introduction.tex
\section{Introduction} 
\label{sec:intro}
\IEEEPARstart{W}{e} focus on the problem of efficiently organizing search over very large trees, a key challenge for motion planning with complex dynamics and costs often encountered in autonomous driving~\citep{dolgov2010path} and aerial vehicles~\citep{uav_path_planning}.
In this paper, we are particularly motivated by the problem of planning for autonomous driving in unstructured, off-road environments where the autonomous vehicle (AV) only has access to a body-centric local map~\citep{han2023model,meng2023}.
This problem in general exists for more challenging aspects of on-road driving as well, which lack rules or structure, such as driving in parking lots, construction zones, and rural areas.

This problem has traditionally been cast as a search over a graph of primitive motions, with many advances in constructing good graphs~\citep{knepper2006high,pivtoraiko2005efficient} and in searching efficiently over them~\citep{ferguson2007field, hedegaard2021discrete}.
A key reason for the efficiency is the notion of \emph{dominance} that arises from the Bellman condition~\citep{bellman_curse}; when a new vertex lands on top of an existing vertex, only the one with a lower cost-to-come survives, and the other is pruned.
Thus, given a branching factor $b$ and path length $d$, far fewer than $b^d$ vertices are practically expanded.
Unfortunately, applications where the environment affects the dynamics of the system, as is the case for off-road autonomy~\citep{han2023model, terrainCNN}, will require repeatedly solving the two-point boundary-value problem (BVP) to build the graph~\citep{damm2023terrain}, which can be too slow for kinodynamic planning for fast moving systems with computational constraints~\citep{SST, Sparse_RRT}.

Dolgov et al.~\citet{dolgov2010path} 
addressed this challenge by proposing the Hybrid A*(\hastar) algorithm, which commits to search over a tree of  motion primitives instead. 
Their key insight was to introduce the  notion of \emph{approximate dominance}, where an expanded tree vertex only had to fall within a grid cell of a previously-searched vertex for dominance to be checked. 
Hybrid A* soon became the de facto standard for planning for unstructured driving~\citep{hastar_real_time_impl,hastar_parking}.
However, the resolution of this grid is hard to tune.
A low resolution over-prunes~(\figref{fig: main_0}), failing to find a path, whereas a high resolution over-expands, resulting in slow planning times~(\figref{fig: main_4}).
Effectively, there is a coupling between vertex discovery and approximate dominance.

Our key insight is that we can reframe search over trees not as rigidly pruning the large tree like \hastar proposes, but as dynamically organizing vertex expansion via an \emph{anytime} planning framework which breaks this coupling. 
Our framework, which we term Incremental Generalized Hybrid A* (\ighastar), enables automatically changing the resolution at which vertex expansions occur.
We show how different variants of \ighastar organize search based on the structural assumptions they make. In addition, we prove that \ighastar always does as well as, if not better than \hastar.
We evaluate \ighastar on planning queries from the Moving AI benchmark~\citep{moving_AI_benchmark} and BeamNG simulator~\citep{beamng_tech}, showing that where an internal baseline -- \ihastar, an optimized version of \hastar that also increases its search resolution--finds the optimal path, all variants of \ighastar find the same path with $\approx6\times$ fewer expansions.
In simulated off-road autonomy experiments using the BeamNG~\citep{beamng_tech} simulator, \ighastar outperforms \ihastar in completion time under a fixed expansion budget when run in a closed-loop fashion with an MPC~\citep{han2023model}, running at $\approx$4~Hz.
We also demonstrate real-time performance ($\approx$3-4~Hz) on a 1/10th scale off-road vehicle sharing compute with MPC and perception on an Nvidia Jetson Orin NX.

%% file: sections/literature.tex
\section{Related work}
\label{sec:related}
Off-road environments often lack structured cues like lane markers or signs, and systems are typically deployed without prior maps~\citep{planetary}, requiring real-time planning in higher-dimensional spaces.
Methods such as \RRTstar~\citep{RRT*} have been applied to off-road planning in kinematic space~\citep{RRT_offroad_kinematic}, while kinodynamic planning has mostly focused on systems with linearizable dynamics~\citep{Kinodynamic_RRT_linearizable_only}, often unsuitable for off-road vehicles that depend heavily on complex terrain interactions~\citep{han2023model, terrainCNN}. 
Recent work~\citep{damm2023terrain} addresses this gap by explicitly constructing graphs with full kinodynamic models.
However, all these approaches require repeatedly solving costly two-point BVPs~\citep{SST, Sparse_RRT}. 
Here, we commit to efficient search over implicit trees generated purely via forward model propagation.

\hastar~\citep{dolgov2010path} is popular for on-road navigation~\citep{hastar_real_time_impl,hastar_parking} relying on the notion of pruning (approximately) \textit{duplicate} states as introduced by Barraquand and Latombe~\citet{barraquand}.
Methods like Sparse-\RRT~\citep{Sparse_RRT} and SST~\citep{SST} also prune states to accelerate search, using r-discs instead of grids. 
Other work~\citep{soft_duplicate, RCS_needle} defines duplicates via similarity of successor validity. 
In general, these methods are sensitive to the resolution (or equivalent hyperparameters) used for duplicate detection. 
Here, we propose a general framework to automatically adapt this hyperparameter on the fly in an anytime fashion, akin to how Anytime \astar-like algorithms~\citep{ara*,ana*,aapex} adjust heuristic weights dynamically.

Methods like Multi-Resolution \astar (\mrastar)~\citep{mrastar, amra*} perform simultaneous searches at multiple resolutions, with discrete graphs defined at multiple spatial resolutions but fundamentally rely on exact state overlap
across graphs for information sharing, which is not guaranteed~(measure 0) with complex system dynamics in continuous space.
In contrast, our approach, as HA*~\citep{dolgov2010path}, commits to searching over the tree generated by the system's dynamics.
Prior approaches~\citep{yahja1998framed, garcia2014gpu} adapt resolution \textit{locally} (such as near obstacles) by relying on domain knowledge. 
Our method changes resolution globally and requires no such priors, making it domain-agnostic.
Our work builds on the notion of the informed set from prior work~\citep{bit*, guild, YiTGSS18}, and is conceptually related to graph densification approaches~\citep{graph_densification}, where the search is performed over a pre-defined set of graphs of increasing density.
However, our trees are not predefined; instead, we generate sub-trees on the fly by forward-propagating the system dynamics.

%% file: sections/Problem.tex
\section{Complexity of searching over a tree}
\label{sec:problem}
We consider the problem of computing a near-optimal path over an implicitly-defined tree $T = (V, E)$, where each edge $(u, v) \in E$ has an associated positive cost given by a function $w : E \rightarrow \mathbb{R}^+$. 
The tree is rooted at a start vertex $v_s \in V$, and the objective is to reach any vertex in a goal set $V_g \subseteq V$ while minimizing the accumulated edge cost.
The tree structure is defined implicitly by a successor function $\Succ$ that returns the set of valid children for any given vertex. 
With a slight abuse of notation, we set $w(\pi)$ to be the cost of a path $\pi = \langle v_1, v_2, \ldots, v_n \rangle$. Namely,
$w(\pi):=\sum_{i=1}^{n-1} w(v_i, v_{i+1})$. Additionally, we define $w(\emptyset) = \infty$ to denote the cost of failure.
We assume access to a heuristic function $h: V \rightarrow \mathbb{R}_{\ge 0}$ that is consistent and admissible, providing an estimate of the remaining cost to the goal.
A baseline algorithm for solving this problem is A* search applied directly to the tree. 
The algorithm maintains a priority queue ($Q_v$) of vertices sorted by $v.f = v.g + h(v)$, where $v.g$ is the cost of the path from $v_s$ to $v$, and $h(v)$ is the heuristic estimate to the goal. 
The algorithm begins by inserting $v_s$ into the queue with $v_s.g = 0$, and repeatedly expands the node with the lowest $f$-value, generating all of its children via $\Succ$, until a goal vertex in $V_g$ is reached. 
As this algorithm considers every child at each expansion, its worst-case complexity is exponential in the depth $d$ of the tree, i.e., $O(b^d)$, where~$b$ is the maximum branching factor. Since no two vertices coincide in the tree, there is no opportunity to prune, and the true complexity worryingly approaches the worst case. This is effectively Alg.~\ref{alg:hastar} without the approximate pruning block highlighted in red.
\vspace{-10pt}

%% file: sections/Method.tex
\section{Hybrid \astar(\hastar)}\label{hastar}
\begin{algorithm}[t]
\small
    \caption{\small Hybrid A* Search on a Tree}
    \label{alg:hastar}
    \textbf{Input:} Start vertex $v_s$, Goal vertex set $V_g$, Resolution $R$ \\
    \textbf{Output:} Least-cost path or $\emptyset$ if FAILURE
    \begin{algorithmic}[1]
        \Function{\hastar}{$v_s, V_g, R$}
        \State $Q_v \gets \{v_s\}$ \Comment{Priority: $v.g + h(v)$} \label{ll:ha_Q_v}
        \While{$Q_v$ $\neq \emptyset$} 
            \State $u \gets Q_v.\textsc{pop()}$ \label{ll:ha_pop}
            \If{$u \in V_g$} 
                \State \Return \textsc{EmitPath}()
            \EndIf
            \ForAll{$v \in \Succ(u)$} \label{line: gha_succ}
                 \State $v.g \gets u.g + w(u, v)$ 
                \textcolor{red}{
                \If{$v.g < \hat{v}(v,R).g$} \label{ll:gha_dominance_check}\Comment{Approx. dominance}
            \State $Q_v.\text{Remove}(\hat{v}(v,R))$
            \State $\hat{v}(v,R) \leftarrow v$ 
            \State $\text{$Q_v$.Insert}(v)$ \label{line: gha_Q_v_insert}
                \EndIf
                }
            \EndFor
        \EndWhile \label{line: gha_inner_end}
        \State \Return $\emptyset$
    \EndFunction
    \end{algorithmic}
\end{algorithm}
 The Hybrid A* algorithm (\hastar, Alg.~\ref{alg:hastar}) was introduced by Dolgov et al.~\citet{dolgov2010path} for solving the motion-planning problem for a kinematic car in urban environments by reducing it to a search over a large tree.
Like the aforementioned baseline algorithm, \hastar also expands vertices in order of increasing $f$-value.
The key difference lies in how it reduces search effort by \textit{approximate dominance} (highlighted in \textcolor{red}{red} in Alg.~\ref{alg:hastar}): 
when two vertices fall into the same \textit{discretized-region} of the continuous space (a grid cell of resolution $R$ in~\citep{dolgov2010path}), they are subject to a dominance check.
The function $\hat{v} : V \times \mathbb{N} \rightarrow V$ maps a vertex $v$ to the vertex that currently dominates $v$'s grid cell at resolution $R$. If $v$ has a lower $g$-value than its dominant vertex~(Alg.~\ref{alg:hastar} Line~\ref{ll:gha_dominance_check}), it becomes the dominant vertex in this grid cell and is inserted into the queue, and the pre-existing dominant is discarded.
By expanding the domain of dominance from a single vertex to a region, approximate pruning can greatly reduce the number of vertices expanded. Unfortunately, selecting the correct resolution can be extremely hard in practice. Fig.~\ref{fig: main} demonstrates that the discretization resolution used by \hastar can be either too low or too high for a given problem.
In the case of off-road kinodynamic planning, the successor vertices may be farther or closer depending on the speed of the parent vertex, and the speed can change with each successor.
The discretization resolution that may seem to work reasonably well (Fig.~\ref{fig: main_2}) in one search query can easily result in anything between not finding a solution (Fig.~\ref {fig: main_0}) to spending too many expansions for a solution (Fig.~\ref {fig: main_4}).
\begin{algorithm}[t]
    \small
    \caption{\small Multi-Resolution Hybrid A* (\ihastar)}
    \label{alg:ham}
    \textbf{Input:} start vertex $v_s$, goal set $V_g$, resolution sequence $\mathcal{R}$\\
    \textbf{Output:} Least-cost path or $\emptyset$ if \textsc{Failure}
    \begin{algorithmic}[1]
        \State $\hat{\pi} \gets \emptyset$
        \For{$i = 0$ to $N$}
            \State $\pi \gets \hastar(v_s, V_g, R_i)$
            \If{$w(\pi) < w(\hat{\pi})$}
                \State $\hat{\pi} \gets \pi$
            \EndIf
        \EndFor
        \State \Return $\hat{\pi}$
    \end{algorithmic}
\end{algorithm}
\vspace{-5pt}
\section{Incremental Generalized Hybrid \astar~(\ighastar)}\label{sec: ighastar}
\subsection{Warmup: \ihastar - Simple multi-resolution \hastar search}\label{subsec: ihastar}
A straightforward extension of \hastar is to restart the forward search at successively finer \textbf{\textit{levels}} of resolution (given by a sequence $\mathcal{R} = [R_0, R_2, \dots, R_N]$), an algorithm which we call \ihastar (Alg.~\ref{alg:ham}). At each \textbf{\textit{iteration}}, \ihastar runs a complete search from scratch at the current resolution level, discarding all previously expanded vertices. In practice, we also branch-and-bound  on the current best cost $w(\hat{\pi})$ to improve efficiency. Specifically, when we pop a vertex $v$ from the queue, we terminate the search if $v.f \geq w(\hat{\pi})$.
While \ihastar can eventually recover the first and best path, it is still computationally inefficient.
States that were expanded and pruned at coarser resolutions are repeatedly rediscovered through fresh expansions at finer levels.
Our key observation is that although pruning is valuable, prematurely pruning vertices can cut off useful vertices forever.
\subsection{The \ighastar Framework}\label{subsec:ighastar_framework}
\begin{figure}[t!]
    \centering
    \begin{subfigure}[t!]{.355\linewidth}
        \centering
        \includegraphics[width=\linewidth]{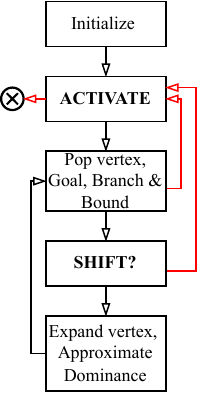}
        \caption{Flowchart}
        \label{fig:ighastar-flow}        
    \end{subfigure}
    \begin{subfigure}[t!]{0.61\linewidth}
        \centering
        \begin{minipage}[t!]{\linewidth}
            \begin{algorithmic}[1]
\small
\Function{Activate}{}
    \ForAll{$v \in Q_v$}
        \If{$\hat{v}(v, R_l) = v$}
            \State $v\text{.Active} \gets \textbf{true}$
        \Else
            \State $v\text{.Active} \gets \textbf{false}$
        \EndIf
    \EndFor
\EndFunction
\Function{Shift}{}
    \State Initialize $H \gets 0$; persists
    \State $u\gets Q_v.\text{Active.}\textsc{peek}()$
    \If{$u.\text{DomLevel} < l$}
        \State $H \gets H + 1$
        \If{$H > \bar{H}$}
            \State $H \gets 0$
            \State $l'\gets u.\text{DomLevel}$
            \State \Return \textbf{true}
        \EndIf
    \EndIf
    \State $l'\gets l + 1$
    \State \Return \textbf{false}
\EndFunction
            \end{algorithmic}
        \end{minipage}
    \caption{\textsc{ACTIVATE} and \textsc{SHIFT} for \ighastar-$\bar{H}$}
    \label{fig:ighastar-activate-shift}
    \end{subfigure}
    \caption{\ighastar decopules vertex dominance and activation. (\subref{fig:ighastar-flow})~\textcolor{red}{Red arrows} denote breaking a loop. (\subref{fig:ighastar-activate-shift}) Instantiations of \shift, \activate for \ighastar-$\bar{H}$, where $\bar{H}$ is a measure of the \textit{hysteresis} in switching to a lower resolution}
    \label{fig:ighastar-all}
    \vspace{-15pt}
\end{figure}
At the heart of Hybrid A* variants lies the interplay between \emph{approximate dominance and activation}.
Only dominant vertices in the forward search are ever allowed to \textit{actively} participate in the search process, meaning dominance controls the tree expansion and therefore vertex discovery.
However, outside of the forward search, this interplay can be broken.
Our key insight is to generalize this framework by \emph{decoupling vertex discovery and dominance}.
This separation gives us a powerful new design tool: rather than relying solely on forward search to explore promising regions, we can \emph{adaptively reuse, resurrect, or even inject} vertices from previous search efforts by activating them based on domain insight or resolution dynamics; a framework which we term \ighastar (Alg.~\ref{alg:igha}).

For this decoupling, we introduce two modular mechanisms, \shift and \activate, and illustrate them through high-level control flow of \ighastar(Fig.~\ref{fig:ighastar-flow}).
\activate decides which vertices participate in the following iteration of the forward search.
The planner activates vertices, enters the forward search, and expands vertices by order of priority while checking whether further search is warranted.
\shift decides what the next search resolution will be and can continue the search or break the loop.
Whenever the loop is broken, \activate again decides which vertices in $Q_v$ will be activated for the next search iteration, and this goes on until $Q_v=\emptyset$.
In the following sections(Sec.~\ref{subsec: information_reuse}, ~\ref{subsec: ighastar-H}), we illustrate how specific instantiations of \shift and \activate result in \ighastar variations, and use Fig.~\ref{fig: R2_experiments} to draw out the distinctions between them.

\begin{algorithm}[t]
\small
\caption{\small Incremental Generalized Hybrid A* (\ighastar)
}
\label{alg:igha}
\textbf{Input:} Start vertex $v_s$, Goal set $V_g$, Resolution sequence $\mathcal{R}$\\
\textbf{Output:} Least-cost path or $\emptyset$ if \textsc{Failure}
\begin{algorithmic}[1]
\State $Q_v \gets \{v_s\}$, $v_s\text{.Active} \gets \textbf{true}$
\State $\hat{\pi} \gets \emptyset$, $l \gets 0$, $l' \gets 0$
\While{$Q_v\neq\emptyset$}\Comment{\textbf{\textit{Iterations}} over the \textbf{\textit{forward search}}}
    \State \activate()
    \While{$Q_v.\text{Active} \neq \emptyset$}\label{ll:igha_search_loop_start}
        \If{$Q_v.\text{Active.}\textsc{peek}().f \geq w(\hat{\pi})$}
            \textbf{break}\label{ll:igha_peek_active}
        \EndIf
        \If{$Q_v.\text{Active.}\textsc{peek}() \in V_g$}\label{ll:igha_goal_check}
            \State $\hat{\pi} \gets \textsc{EmitPath}()$
            \State \textbf{break}
        \EndIf
        \If{\textsc{Shift}()}
            \textbf{break}\Comment{Internally update $l'$}
        \EndIf
        \State $u \gets Q_v.\text{Active.}\textsc{pop}()$\label{ll:igha_popactive}
        \ForAll{$v \in \Succ(u)$}
            \If{$v.g < \hat{v}(v, R_l).g$}
                \State $\hat{v}(v, R_l)\text{.Active} \gets \textbf{false}$\label{ll:igha-dom-deactivate}
                \State $\hat{v}(v, R_l) \gets v$
                \State $v\text{.Active} \gets \textbf{true}$
            \Else
                \State $v\text{.Active} \gets \textbf{false}$\label{ll:igha-deactivate}
            \EndIf
            \State $Q_v.\textsc{Insert}(v)$\label{ll:igha_Q_v_save}
        \EndFor
    \EndWhile
    \State $\textsc{Bound}(w(\hat{\pi}))$\Comment{$\forall v\in Q_v$ Remove $v:v.f>w(\hat{\pi})$}\label{ll:igha_bound}
    \State $\textsc{Project}(R_l)$\Comment{$l \gets l'$, update $R_l$, and recompute $\hat{v}$}\label{ll:igha_project}
\EndWhile
\State \Return $\hat{\pi}$
\end{algorithmic}
\end{algorithm}
\subsection{\ighastar - Multi-resolution search with information reuse}\label{subsec: information_reuse}
In this section, we dive into the details of our algorithm
and design a straightforward but effective instantiation of \ighastar (Alg.~\ref{alg:igha}).
Here, during forward search~(Line~\ref{ll:igha_search_loop_start}) dominant vertices are marked as active, non-dominant as inactive, and both are saved in $Q_v$~(Lines~\ref{ll:igha-dom-deactivate},~\ref{ll:igha-deactivate}, and~\ref{ll:igha_Q_v_save}) for potential use in future resolutions with the invariant that only active vertices from $Q_v$ can be expanded~(Line~\ref{ll:igha_popactive}).
Note that $Q_v.\text{Active}$ refers to the partition of $Q_v$ containing active vertices.
The forward search concludes if either we've run out of vertices to expand, the branch and bound check fails, or a better path has been found (Lines~\ref{ll:igha_search_loop_start}, ~\ref{ll:igha_peek_active},~\ref{ll:igha_goal_check} resp.).
Then, we recheck the branch and bound, \emph{project} all vertices to a finer resolution~(Lines~\ref{ll:igha_bound}, ~\ref{ll:igha_project} resp.), activate vertices newly dominant at this resolution (Fig.~\ref{fig:ighastar-activate-shift}, \activate), and continue the search seamlessly like before at the finer resolution.
Monotonically improving paths are emitted whenever they are discovered, as with \ihastar, while requiring fewer expansions through vertex reuse.
\vspace{-10pt}
\subsection{\ighastar-$\bar{H}$ - Dynamically \shift resolution up and down}\label{subsec: ighastar-H}
\begin{figure*}[!htb]
\centering
\begin{subfigure}{0.19\linewidth}
  \includegraphics[width=\linewidth]{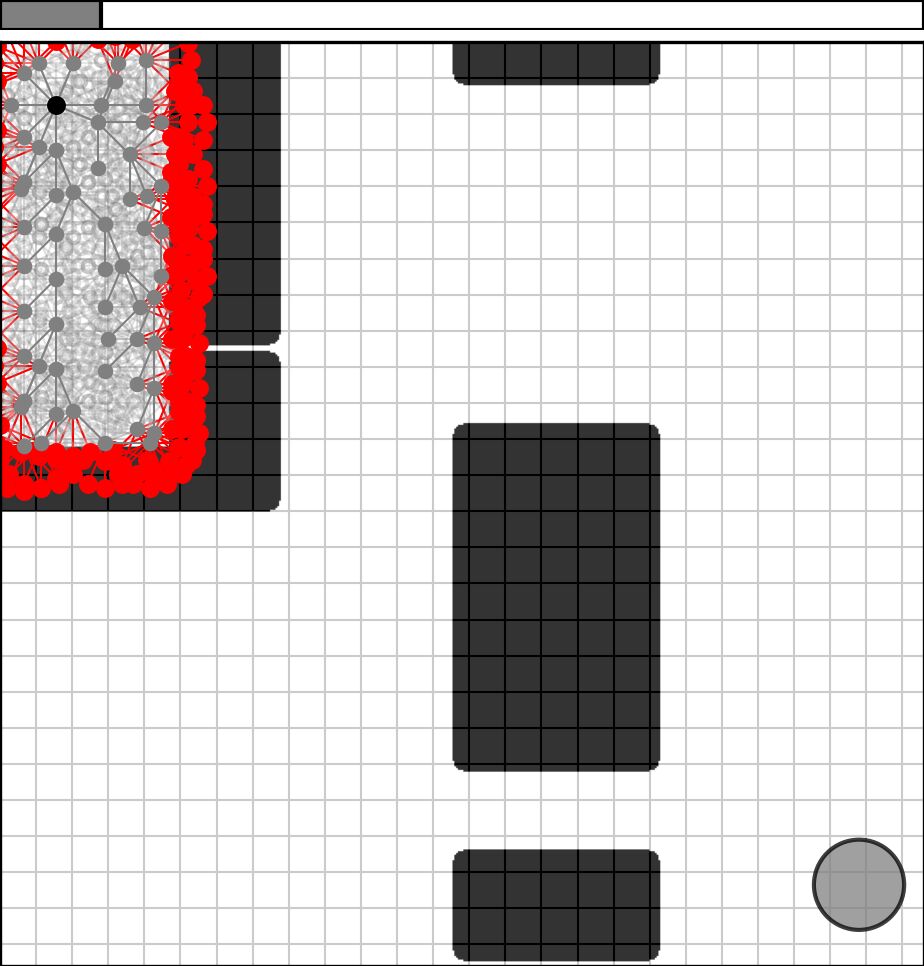}
  \caption{SB, \hastar}
  \label{fig:single_bn_example}
\end{subfigure}%
\hspace{0.1pt}
\begin{subfigure}{0.19\linewidth}
  \includegraphics[width=\linewidth]{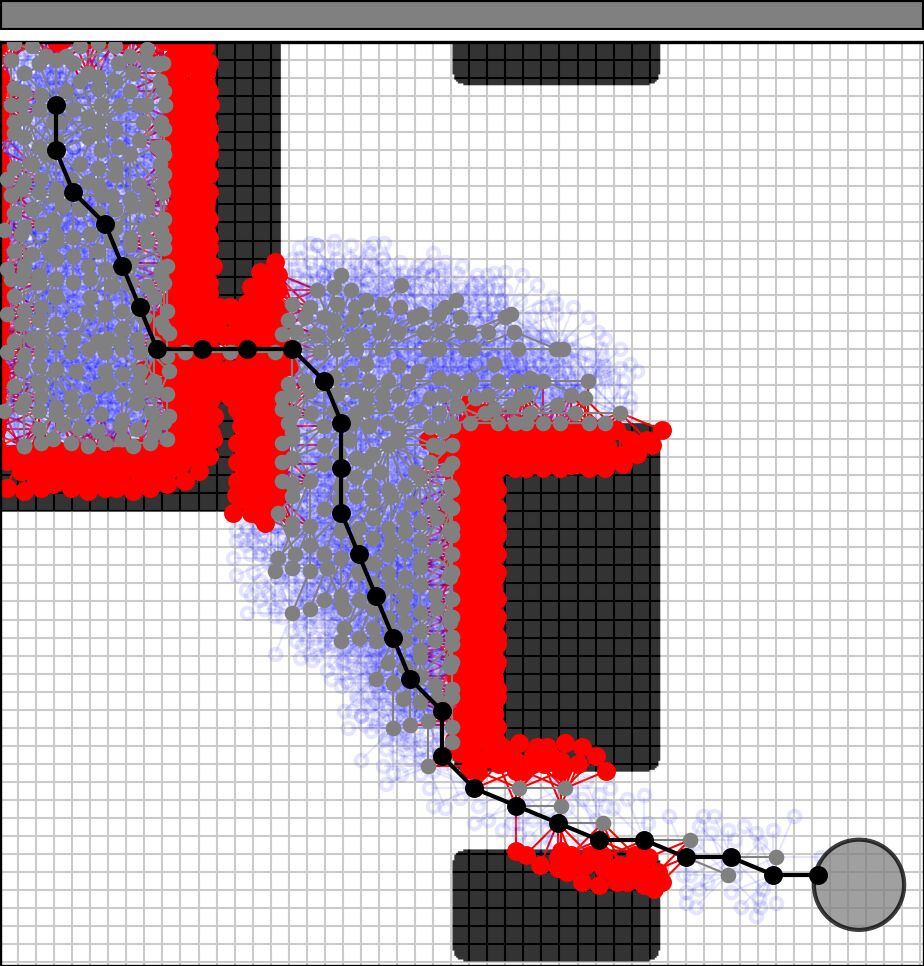}
  \caption{SB, \drk}
  \label{fig:single_bn_drk}
\end{subfigure}%
\hspace{0.1pt}
\begin{subfigure}{0.19\linewidth}
  \includegraphics[width=\linewidth]{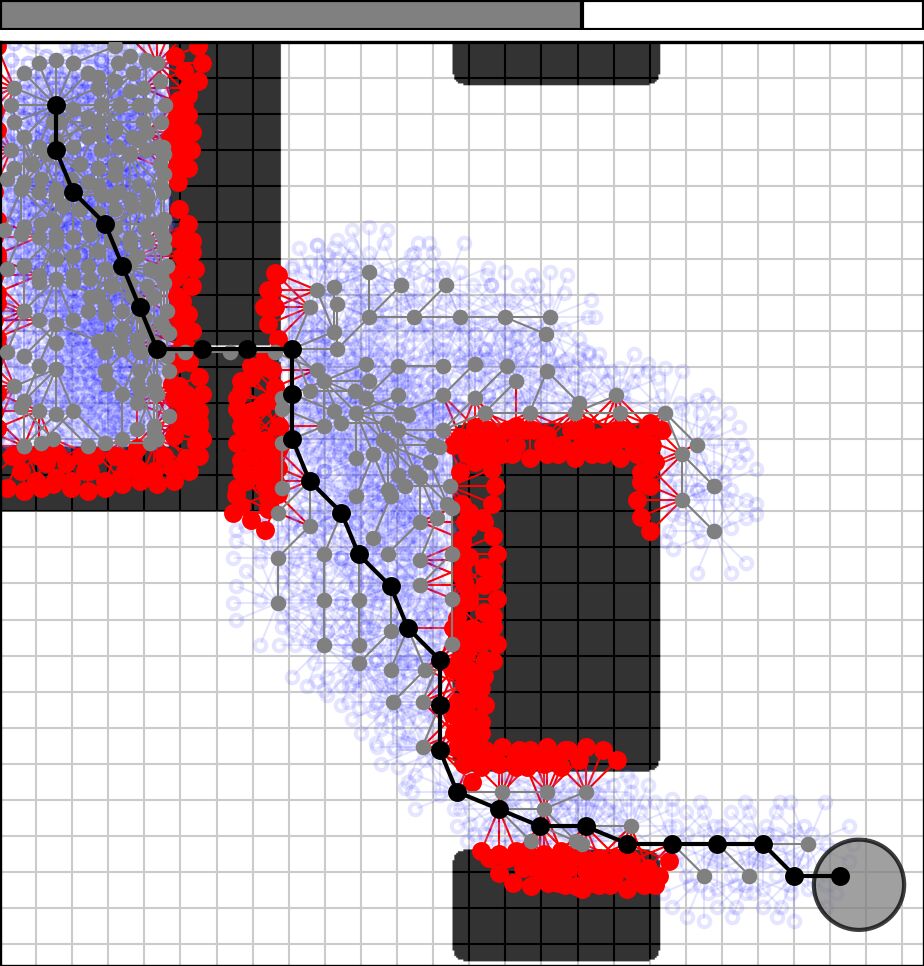}
  \caption{SB, \ighastar-50}
  \label{fig:single_bn_50}
\end{subfigure}%
\hspace{0.1pt}
\begin{subfigure}{0.19\linewidth}
  \includegraphics[width=\linewidth]{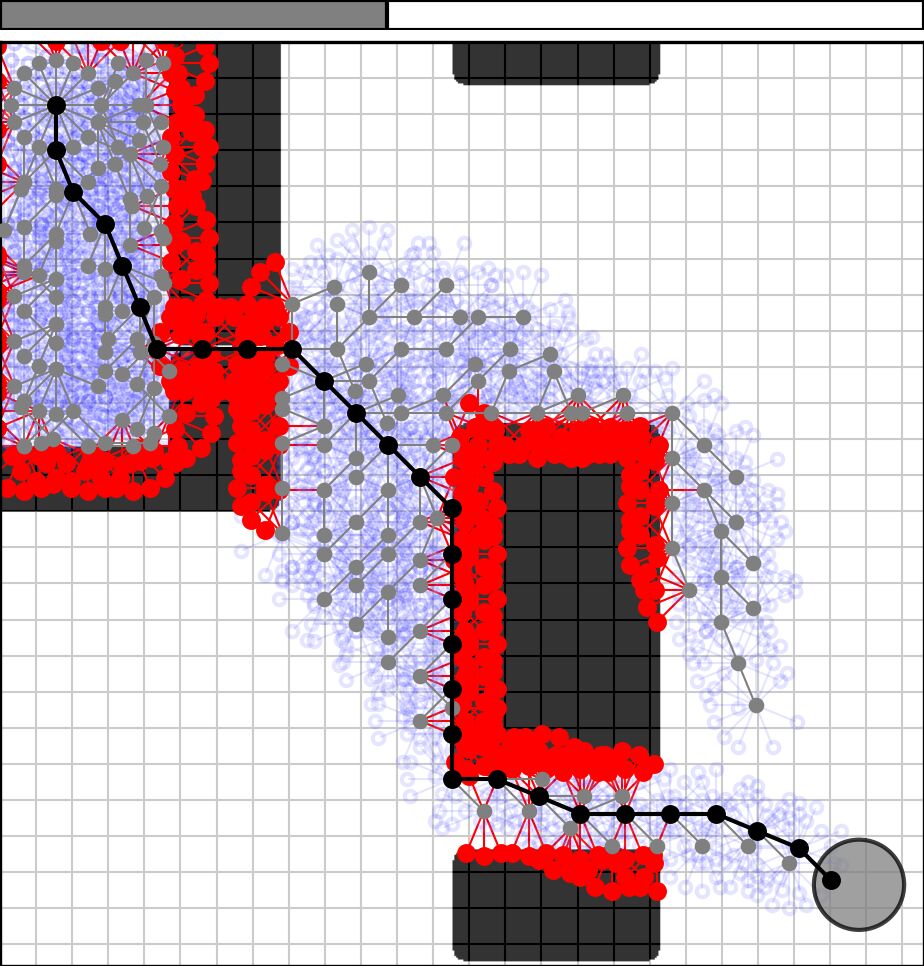}
  \caption{SB, \dsrk}
  \label{fig:single_bn_dsrk}
\end{subfigure}%
\hspace{0.1pt}
\begin{subfigure}{0.20\linewidth}
  \includegraphics[width=\linewidth]{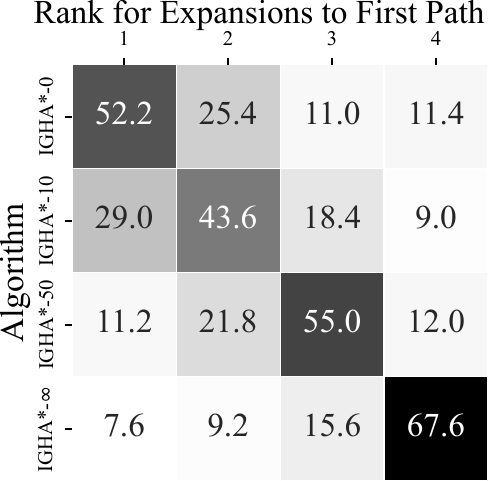}
  \caption{SB Rank Plot}
  \label{fig:single_bn_rankplot}
\end{subfigure}%
\\
\begin{subfigure}{0.19\linewidth}
  \centering
  \includegraphics[width=\linewidth]{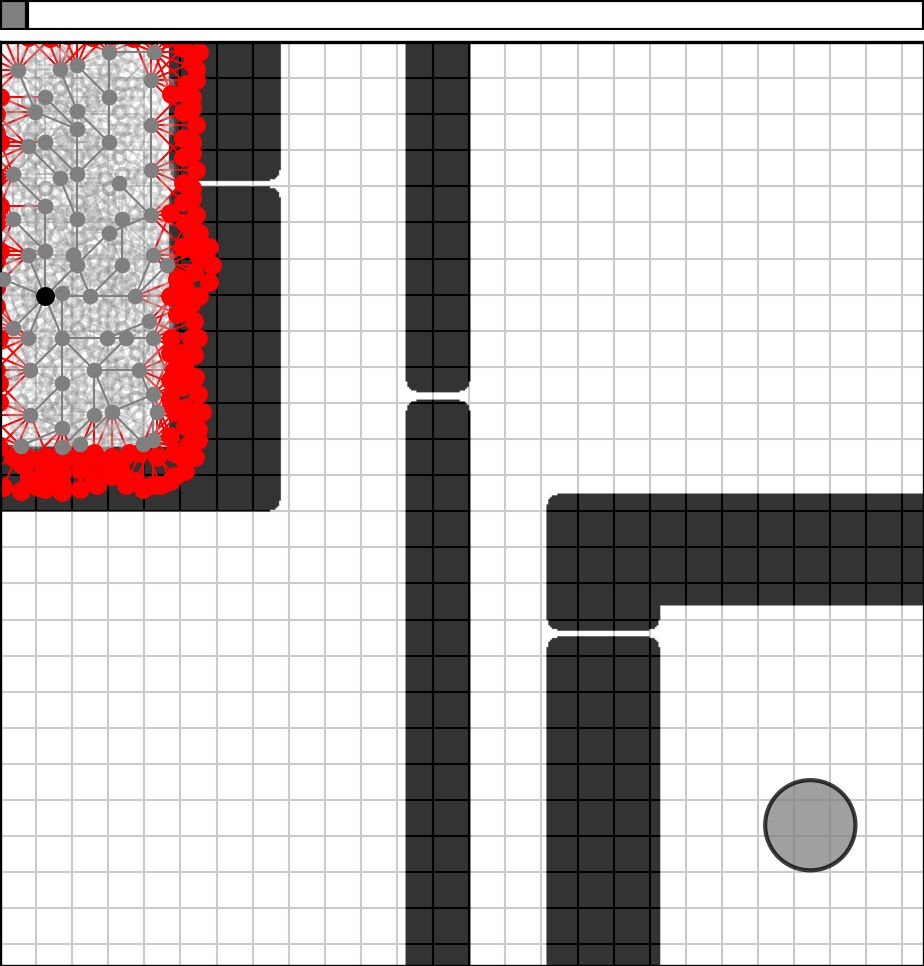}
  \caption{MB \hastar}
  \label{fig:multi_bn_example}
\end{subfigure}
\begin{subfigure}{0.19\linewidth}
  \centering
  \includegraphics[width=\linewidth]{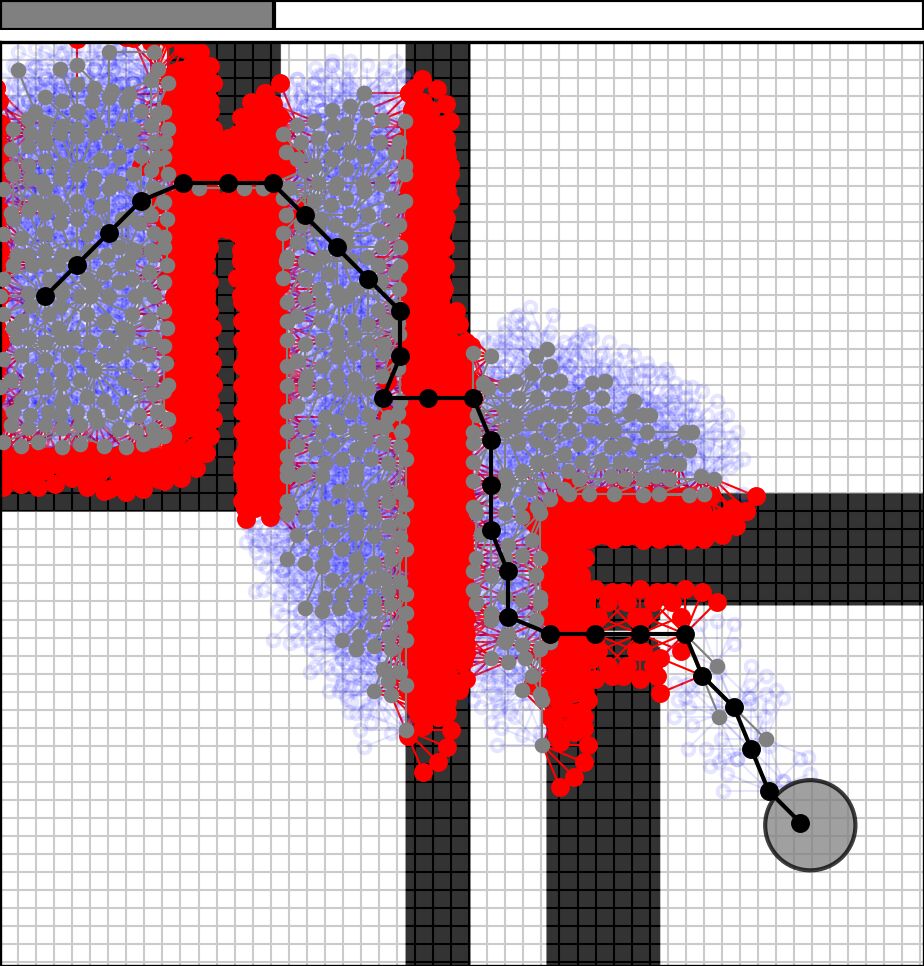}
  \caption{MB \drk}
  \label{fig:multi_bn_drk}
\end{subfigure}
\begin{subfigure}{0.19\linewidth}
  \centering
  \includegraphics[width=\linewidth]{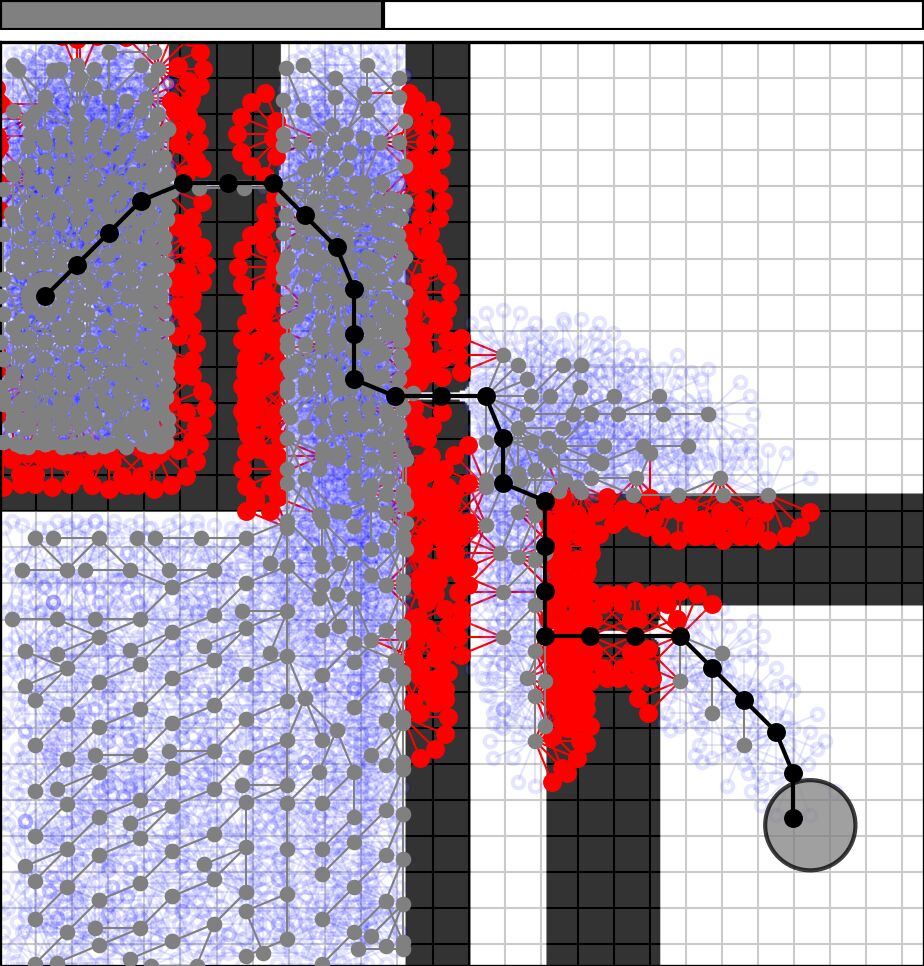}
  \caption{MB \ighastar-50}
  \label{fig:multi_bn_50}
\end{subfigure}
\begin{subfigure}{0.19\linewidth}
  \centering
  \includegraphics[width=\linewidth]{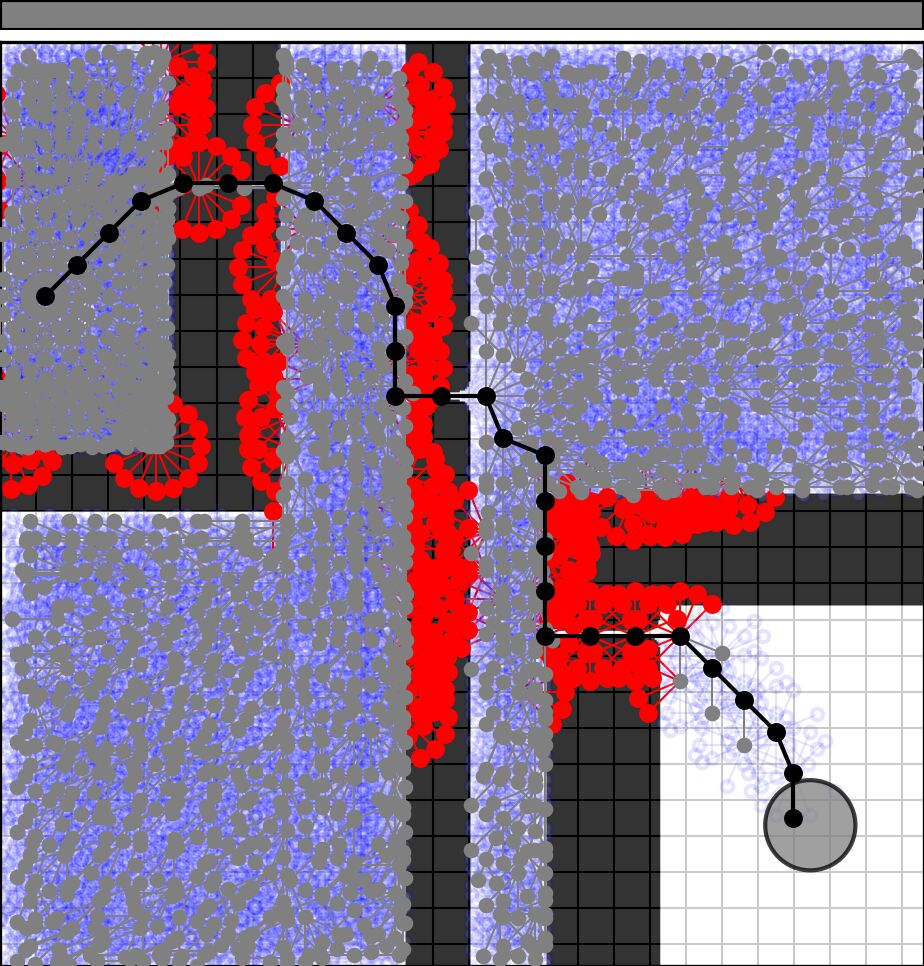}
  \caption{MB \dsrk}
  \label{fig:multi_bn_dsrk}
\end{subfigure}
\begin{subfigure}{0.20\linewidth}
  \centering
  \includegraphics[width=\linewidth]{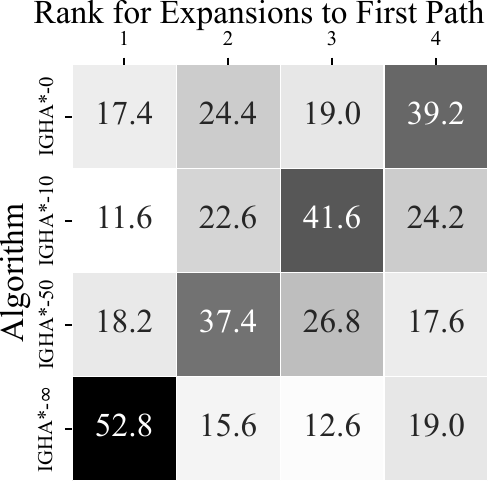}
  \caption{MB Rank Plot}
  \label{fig:multi_bn_rankplot}
\end{subfigure}
\caption{
Here, the resolution is too low for \hastar to find a solution for either the Single-Bottleneck~(\textbf{SB}) scenario~(\subref{fig:single_bn_example}, or the Multi-Bottleneck~(\textbf{MB}) scenario~\subref{fig:multi_bn_example}).
In contrast to Fig.~\ref{fig: main}, for figures corresponding to \ighastar, translucent blue vertices are \textit{deactivated}.
\ighastar variants like \drk only increase resolution and spend more expansions after escaping a bottleneck~(\subref{fig:single_bn_drk}) when a lower resolution could suffice \dsrk(\subref{fig:single_bn_dsrk}), resulting in fewer expansions to the first path for \dsrk(\subref{fig:single_bn_rankplot}). 
However, variants like \dsrk can prematurely switch back in situations where staying at the higher resolution is necessary and spend extra expansions~(\subref{fig:multi_bn_dsrk}) by comparison to \drk(\subref{fig:multi_bn_drk}).
In both scenarios, an intermediate value for $\bar{H}$ can provide intermediate behavior; doing better than \drk in SB~(\subref{fig:single_bn_50}), and better than \dsrk in MB~(\subref{fig:multi_bn_50}).
This interpolation is also visible through the rank plots corresponding to either scenario.
In the \textbf{rank plot}, cells show the percentage of times (darker $=$ higher) a planner achieved a rank for first-path; lower ranks (left) = faster solutions.
}
\label{fig: R2_experiments}
\vspace{-20pt}
\end{figure*}

We have deliberately skipped over \shift until now;
although monotonically increasing resolution is often a fine strategy, consider a scenario where a higher resolution was only needed to pass through a \textit{single bottleneck} (SB)~(Fig.~\ref{fig:single_bn_example}).
Once vertices are through the bottleneck, we would observe dominance not only at the current but also lower resolutions.
Staying at a higher resolution when a lower one would suffice results in an influx of vertices that would have been deactivated otherwise, creating more work.
Instead of committing to the higher resolution (Fig.~\ref{fig:single_bn_drk}), this information of \textit{dominance at a lower resolution} can be used as an indicator to \shift to a lower resolution as soon as we pop an active vertex that indicates this~(Fig.~\ref{fig:single_bn_dsrk}) and save on expansions.

However, if there are \textit{multiple bottlenecks} (MB) to pass through that the search process has not encountered yet~(Fig.~\ref{fig:multi_bn_example}), switching back at the first sign of low resolution dominance~(Fig.~\ref{fig:multi_bn_dsrk}) can also work against us; we switch to a low resolution, cannot anymore fit through the next bottleneck, expand into the empty regions around the bottleneck, wasting effort, and once we've expanded as much as we can, finally switch to a higher resolution.
Predicting imminent bottlenecks in general configuration spaces may not always be possible.
While we may have some intuition based on domain knowledge, we don't \textit{exactly} know which situation we are in a priori. 
A safe bet is to incorporate some \textit{hysteresis} until we've seen some evidence before switching to a lower resolution.
This is the essence of \ighastar-$\bar{H}$'s \shift function (Fig.~\ref{fig:ighastar-activate-shift}) --- the hysteresis threshold $\bar{H}$
allows us to interpolate between the two extremes of $\bar{H} = \infty$, which monotonically increases resolution, and $\bar{H} = 0$, which switches to a lower resolution as soon as it can (See Fig.~\ref{fig:multi_bn_rankplot},~\ref{fig:single_bn_rankplot} and caption), effectively allowing us to encode a structural assumption into the algorithm through \shift and \activate.
Note that we omit certain implementation details for simplicity, see \textit{Implementation} in Sec.~\ref{sec:discuss}.

\vspace{-10pt}
\subsection{Questions to address}
Our work motivates the following key questions:
\begin{enumerate}
    \item Are the \shift and \activate methods in Fig.~\ref{fig:ighastar-activate-shift} the only possible implementations? In Sec.~\ref{sec:theory} we discuss the theoretical properties of \activate and \shift that ensure convergence.
    \item Have we replaced one parameter ($R$ in \hastar) with another ($\bar{H}$ in \ighastar-$\bar{H}$)? We show in Sec.~\ref{sec:experiments} that although certain $\bar{H}$ outperform others, \emph{all} of them significantly outperform \ihastar. Thus, if one is allergic to hyperparameters, one can safely set $\bar{H} = \infty$ and ignore \shift.
    \item How does \ighastar-$\bar{H}$ compare to \ihastar in practice for off-road autonomy? We detail these experiments in Sec.~\ref{subsec: closed_loop_exp}).
\end{enumerate}

\section{Theoretical Analysis}\label{sec:theory}
\begin{definition}[\shift]\label{shift_def}
A \shift method dictates the search's resolution in the following iteration of the forward search.
A \shift method can terminate the forward search by returning \textit{false} after the first iteration and after at least one vertex from $Q_v.\text{Active}$ has been expanded in the current iteration.
We term a \shift method \textbf{\textit{naive}} if it only uses information in $Q_v$.
\end{definition}
\begin{definition}[\activate]\label{activate_def}
An \activate method sets the active/inactive status of the vertices in $Q_v$ after the first iteration of the forward search.
We term an \activate method \textbf{\textit{naive}} if it only uses information in $Q_v$.
\end{definition}
\begin{definition}[Rule]\label{rule_def}
A combination of \shift and \activate methods is considered a rule. 
A rule $R$ initializes the forward search of \ighastar, and results in the generation of the sub-trees $T^R_i=(V^R_i,E^R_i),~i\in \mathbb{Z}_{\geq0}$, where $i$ represents the iteration.
Two rules $A,~B$
are considered \textbf{distinct} if for at least one planning query, for at least one iteration $i,~ T^A_i\neq T^B_i$.
Combining a naive \shift and a naive \activate method results in a \textbf{\textit{naive}} rule.
A \textbf{\textit{practical}} rule, for any iteration, does not activate the entire $Q_v$.
A valid rule is such that $|Q_v.\text{Active}|\geq1$ before starting forward search when $Q_v\neq\emptyset$.
\end{definition}
\textbf{Assumptions}:
\begin{enumerate}[label=A\arabic*]
    \item The heuristic $h$  is admissible.
    I.e., $\forall v\in V,~h(v) \leq h^{*}(v)$, where $h^{*}(v)$ is the lowest cost from $v$ to $V_g$.
    \label{heuristic_assumption}
    \item The max. branching factor at all vertices is $b$ with $b<\infty$.\label{tree_assumption}
    \item $w:E\rightarrow\mathbb{R}_{\geq \epsilon}, \epsilon > 0$.\label{positive_edge_assumption}
\end{enumerate}
\begin{observation}\label{same_first}
    For a given planning query and starting resolution, \hastar and \ighastar expand the same vertices at iteration 0 for any rule.
\end{observation}
\begin{observation}
    \label{monotonic_improvement}
    $w(\hat{\pi})$ strictly decreases with \ighastar iterations for any rule when assumption ~\ref{heuristic_assumption} holds.
\end{observation}
\begin{observation}
    \label{atleast_one_vertex}
    From Def.~\ref{shift_def},~\ref{rule_def}, at least one vertex is expanded each iteration as long as $Q_v\neq \emptyset$. 
\end{observation}
\begin{theorem}\label{Termination}
    When at least one solution to a planning query has been found, \ighastar will terminate with finite expansions
    when assumptions \ref{heuristic_assumption}-\ref{positive_edge_assumption} hold.
\end{theorem}
\textbf{Proof:}
Suppose the first path is found at iteration $i$ with cost $w(\hat{\pi})_i$.
For any iteration $j>i$, given assumption~\ref{heuristic_assumption} and the invariant $Q_v.\text{Active.}\textsc{peek}().f~<~w(\hat{\pi})_i$, let $V':\{v\in V|v.g <w(\hat{\pi})_i\}$ represent the set of all vertices that $Q_v$ could have.
Given ~\ref{tree_assumption},~\ref{positive_edge_assumption}, the maximum possible depth~(from $v_s$) is $d<w(\hat{\pi})_i/\epsilon$ and so $|V'|\leq O(b^{d})$, meaning $|V'|$ is upper bounded by a finite number.
Following Obs.~\ref{atleast_one_vertex}, the algorithm must terminate within finite expansions.
$\square$
\begin{observation}\label{best_path}
    From Obs.~\ref{monotonic_improvement},~\ref{atleast_one_vertex}, and Thm.~\ref{Termination}, when at least one solution to a planning query has been found, and assumptions \ref{heuristic_assumption}- \ref{positive_edge_assumption} hold, \ighastar will find the best-cost path in $T=(V,E)$ at or before termination.
\end{observation}
\vspace{-5pt}
Discretization can cause different trees to be generated depending on which vertices in $Q_v$ are expanded.
No naive rule reliably finds the first or best path with minimal expansions, as $Q_v$ lacks information about which subset yields which tree.

\begin{theorem}\label{no_fastest_first}
For any \textbf{naive} and \textbf{practical} rule $A$ 
there exists another \textbf{distinct} naive and practical rule $B$
for which there exists a scenario where 
$B$ finds the first solution with fewer expansions than $A$.
\end{theorem}

\textbf{Proof:}
Throughout the proof and with a slight abuse of notation, we denote \ighastar-$R$ as the instantiation of \ighastar using rule~$R$.
We use $\xi^R_i$ to represent the number of expansions for \ighastar-$R$ \textit{at} iteration $i$, and set $\Xi^R_i = \sum^{k=i}_{k=0}\xi^R_k$. Finally, we use  $\xi^R_i(u,v),~u,v\in V^R_i$ to represent expansions used to reach~$v$ from $u$ at iteration $i$.
Given some scenario, we call vertex $v_b \in V$ a \emph{bottleneck vertex} if  $v_b$ belongs to every path between $v_s$ and some vertex in~$V_g$ and that $v_b$ is one expansion away from $V_g$.
Now consider a scenario containing a bottleneck vertex~$v_b$ where \ighastar-$A$ does not find the solution at iterations~$0$ and~$1$ and does not unfreeze $v_b$ -- possible as $Q_v$ does not contain information about which vertex will lead to $V_g$.
As $V_g$ is one expansion away from~$v_b$, we have that $\xi^R(v_b,V_g)=1$ for any rule $R$ at any iteration.
Define rule $B$ using the same \shift method as rule $A$ but an \activate method that activates only $v_b \in Q^A_v, v_b \notin Q^A_v\text{.Active}$ at the beginning of iteration 1.
Such a rule is possible as rule $A$ is naive and practical, and $B$ is distinct from $A$.

At iteration 1, \ighastar-$B$ finds a solution, $\Xi^B_1 =\xi^B_0+1$.
At iteration $j>1$, \ighastar-$A$ finds a solution.
$\Xi^A_j = \xi^A_0 + \xi^A_1\dots \xi^A_j(v_s,v_b) + 1$.
Following  Obs.~\ref{same_first}, and~\ref{atleast_one_vertex} we have that $\xi^A_0 = \xi^B_0$, and
that $\xi^A_1+\dots \xi^A_j(v_s,v_b)\geq1$.
Therefore $\Xi^A_j \geq \Xi^B_1 + 1$ and thus $\Xi^A_j > \Xi^B_1$.
$\square$
\begin{corollary}\label{no_fastest_best}
For any \textbf{naive} and \textbf{practical} rule $A$ 
there exists another \textbf{distinct} naive and practical rule $B$
for which there exists a scenario where 
$B$ finds the best solution with fewer expansions than $A$.
\end{corollary}
\textbf{Proof:}
In Thm.~\ref{no_fastest_first}, 
it is possible to have a scenario where the first solution is the best solution. Then, using the same mechanism as in Thm.~\ref{no_fastest_first}, the Corollary can be proven.$\square$

%% file: sections/Experiments.tex
\section{Experiments and Results~}\label{sec:experiments}

\begin{figure}[!htb]
\centering
\begin{subfigure}{0.24\linewidth}
  \centering
  \includegraphics[width=\linewidth]{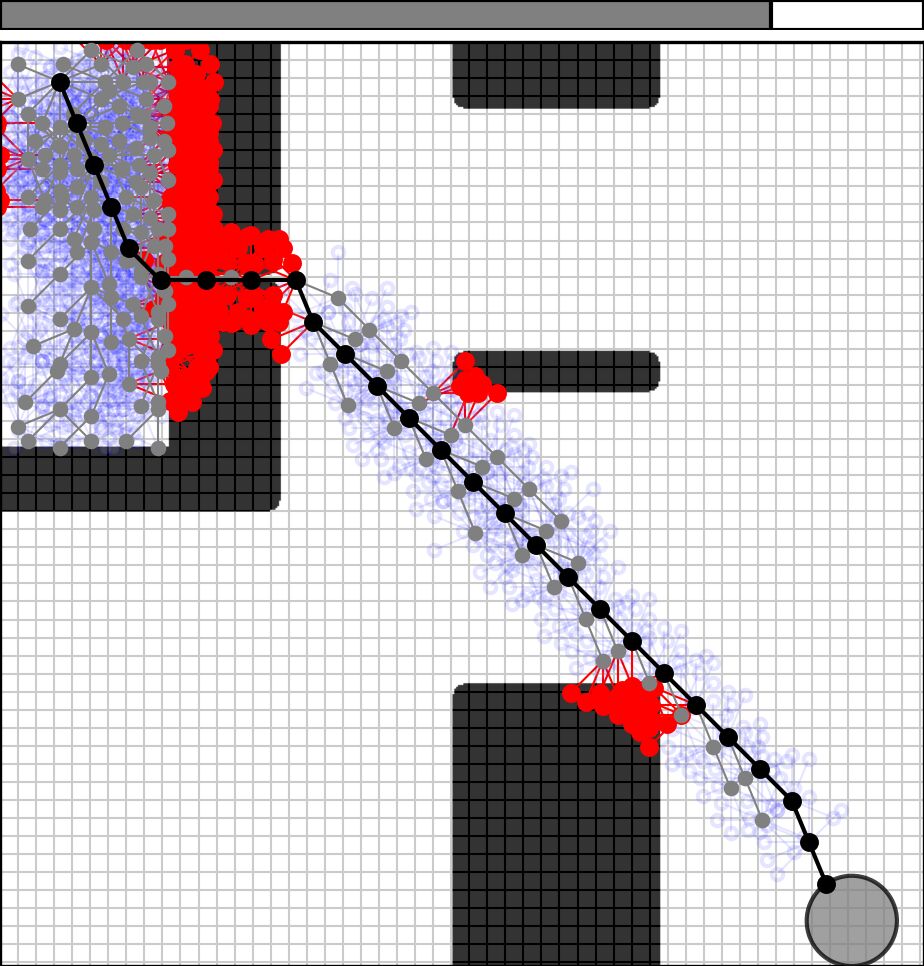}
  \caption{SB \drk}
  \label{fig:single_bn_drk_counter_example}
\end{subfigure}%
\hspace{0.1pt}
\begin{subfigure}{0.24\linewidth}
  \centering
  \includegraphics[width=\linewidth]{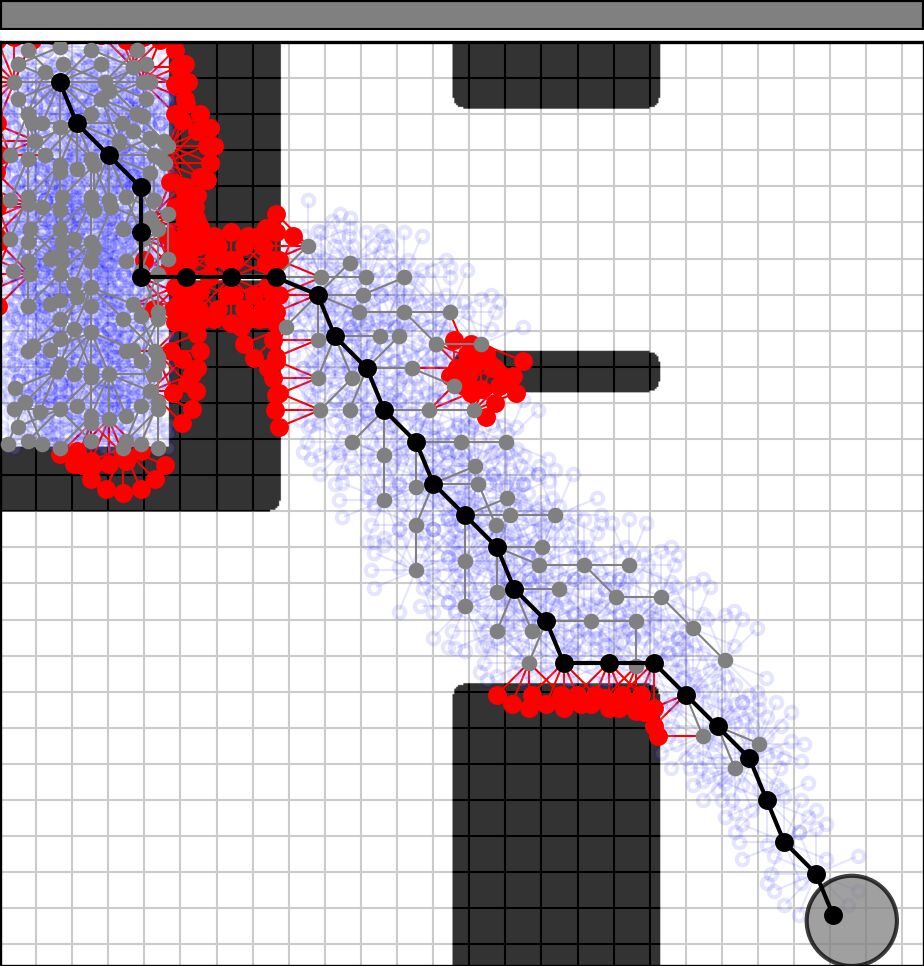}
  \caption{SB \dsrk}
  \label{fig:single_bn_dsrk_counter_example}
\end{subfigure}%
\hspace{0.1pt}
\begin{subfigure}{0.24\linewidth}
  \centering
  \includegraphics[width=\linewidth]{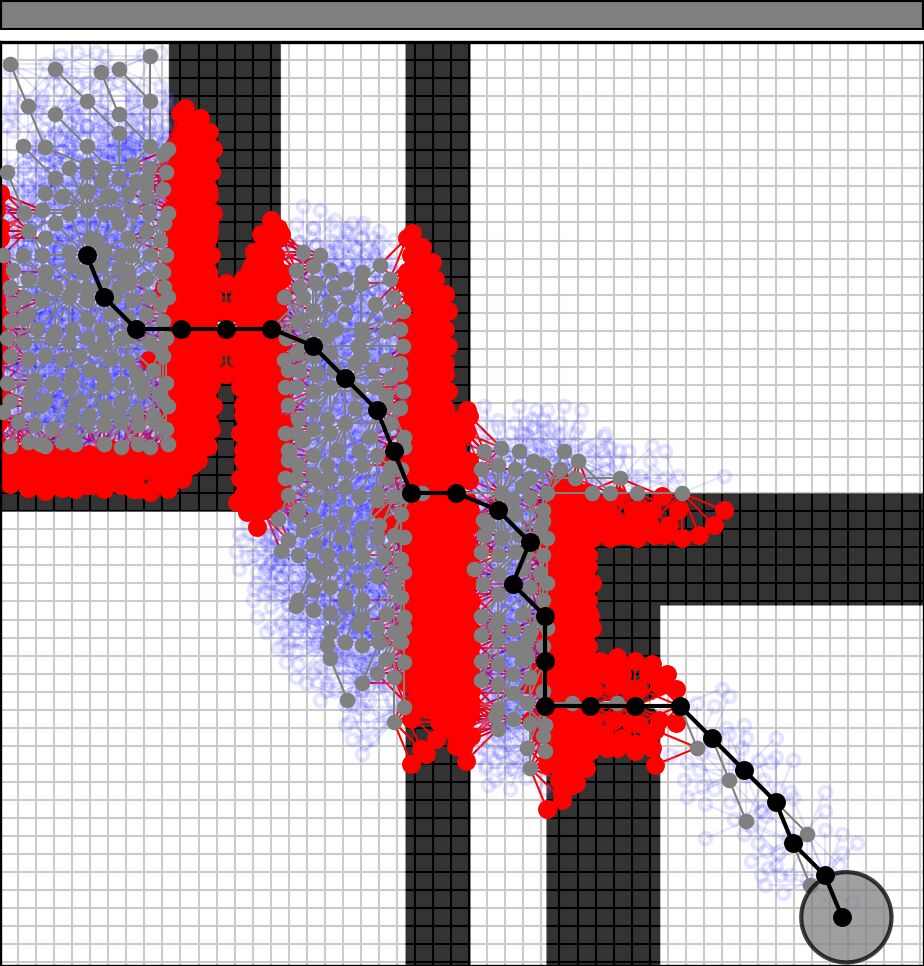}
  \caption{MB \drk}
  \label{fig:multi_bn_drk_counter_example}
\end{subfigure}%
\hspace{0.1pt}
\begin{subfigure}{0.24\linewidth}
  \centering
  \includegraphics[width=\linewidth]{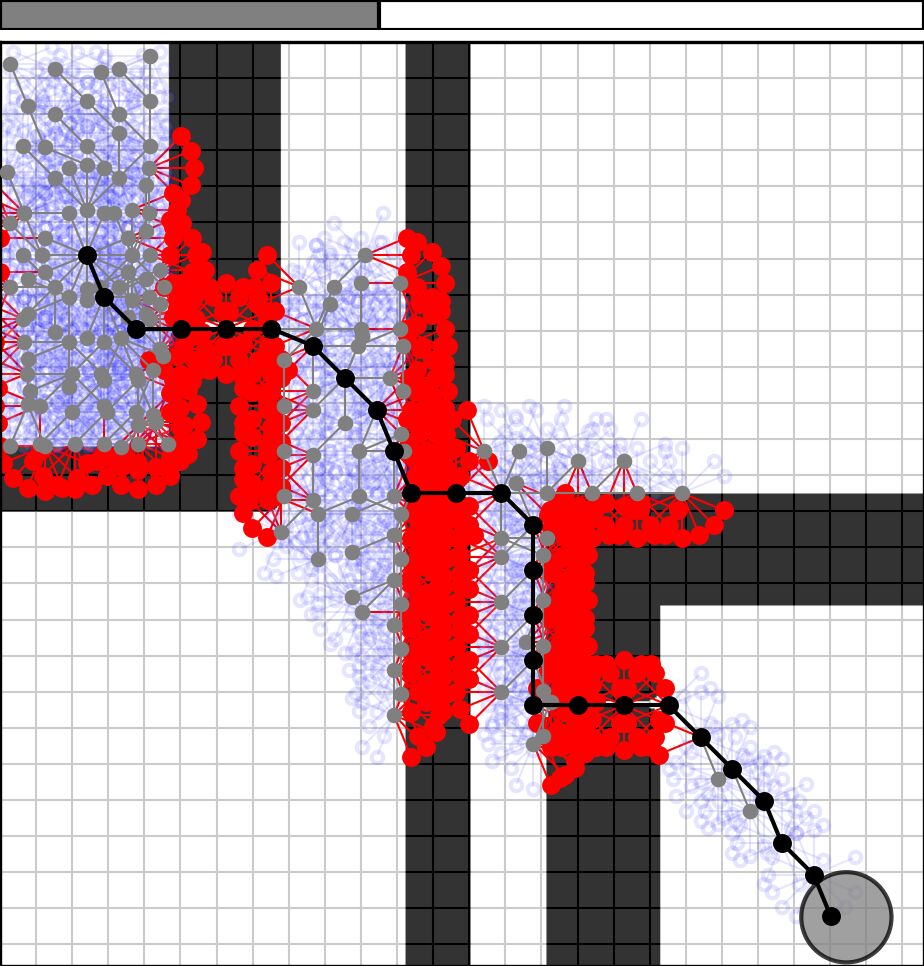}
  \caption{MB \dsrk}
  \label{fig:multi_bn_dsrk_counter_example}
\end{subfigure}%
\caption{
In some queries of the SB environment, \dsrk can have more expansions than \drk (Fig.~\ref{fig:single_bn_rankplot}). Examples~\subref{fig:single_bn_drk_counter_example} and~\subref{fig:single_bn_dsrk_counter_example} show that this happens when the low resolution of \dsrk prevents it from finding a \textit{straight-line} path to the goal.
In some queries of the SB environment, \drk can have more expansions than \dsrk~(\subref{fig:multi_bn_rankplot}).
Examples~\subref{fig:multi_bn_drk_counter_example} and~\subref{fig:multi_bn_dsrk_counter_example} show that this can happen when the tree produced by \dsrk contains a path at a lower resolution that passes through the two other gaps.
}
\label{fig: R2_experiments_counter_Examples}
\vspace{-10pt}
\end{figure}
In this section, \ighastar and \ighastar-$\bar{H}$~(Sec.~\ref{subsec: ighastar-H}) are used interchangeably.
We use \ihastar as a baseline~(Sec.~\ref{subsec: ihastar}).
We use 95$\%$ confidence intervals.
\subsection{$\mathbb{R}^2$ Environments}\label{subsec: R2_exp}
To show that the $\bar{H}$ discussed in Sec.~\ref {subsec: ighastar-H} encodes a structural assumption into \ighastar, we evaluate on 500 automatically generated $\mathbb{R}^2$ environments, with Fig.~\ref{fig:single_bn_example} showing one of the queries for the \textit{Single-Bottleneck} environment~(SB), and Fig.~\ref{fig:multi_bn_example} showing one from the \textit{Multi-Bottleneck} environments~(MB).
We vary the location of the gaps (whitespace) in the walls (black) and vary $v_s$ and $V_g$ in the top left and bottom right corners of the map, respectively.
The planning queries are generated such that \hastar would not find the solution at the starting resolution.
We compare \drk against \dsrk.
We evaluate the total expansions needed to find the first path.

\textbf{Hypotheses:}

\textbf{H1:} In SB, \dsrk finds the first solution with fewer expansions than \drk more often than not.

\textbf{H2:} In MB, \drk finds the first solution with fewer expansions than \dsrk more often than not.

\begin{table}[!htb]
    \centering
    \begin{tabularx}{\linewidth}{lXX}
        \toprule
        \textbf{Comparison} & \textbf{Ratio} & \textbf{Speed Up} \\
        \midrule
        MB \drk/\dsrk  & $0.70 \pm 0.04$ & $3.52 \pm 0.26$ \\
        SB \dsrk/\drk & $0.84 \pm 0.03$ & $1.92 \pm 0.07$ \\
        \midrule
        Kinematic \drk/\dsrk  & $0.67 \pm 0.06$ & $4.10 \pm 0.63$ \\
        Kinodynamic \dsrk/\drk & $0.60 \pm 0.06$ & $3.17 \pm 0.52$ \\
        \bottomrule
    \end{tabularx}
    \caption{
    The \textbf{Ratio} is the ratio of the total queries in which A beats B (A/B) to the first solution.
    The \textbf{Speed Up} shows how much faster A is relative to B \textit{when} A beats B.
    }
    \label{tab:H1_H2_table}
    \vspace{-10pt}
\end{table}
\textbf{Results:}
Table~\ref{tab:H1_H2_table} shows data in the first and second rows that \textbf{confirms hypotheses H1 and H2}, showing that when one beats the other, there is a $2-3\times$ speedup.
We have shown examples of these scenarios in Fig.~\ref{fig: R2_experiments}, showing additionally how intermediate values of $\bar{H}$ operate, and we show counterexamples and how they come up in Fig.~\ref{fig: R2_experiments_counter_Examples}.
\begin{figure}[t]
\begin{subfigure}{.448\linewidth}
  \centering
  \includegraphics[width=\linewidth]{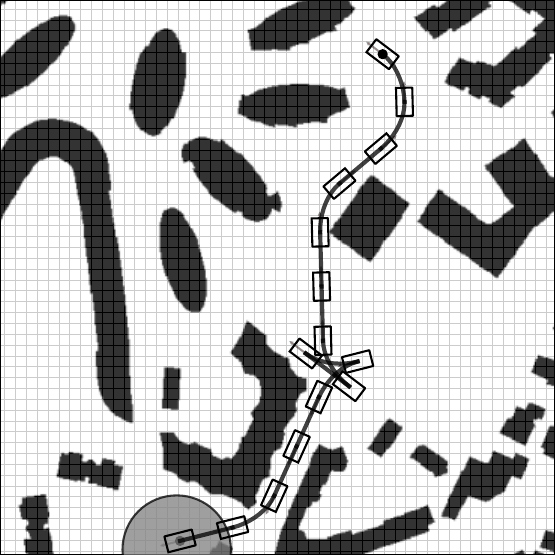}
  \caption{Ubran environments~\citep{moving_AI_benchmark}}
  \label{fig:kinematic_example}
\end{subfigure}%
\hspace{0.5pt}
\begin{subfigure}{0.54\linewidth}
  \centering
  \includegraphics[width=\linewidth]{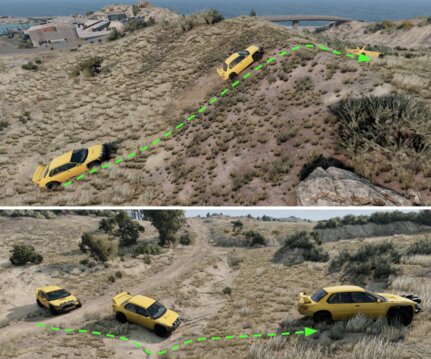}
  \caption{Off road environments~\citep{beamng_tech}}
  \label{fig:off_road_exp}
\end{subfigure}
\caption{
Notice that in Table ~\ref{tab:H1_H2_table}, \drk beats \dsrk in the kinematic environments and \dsrk beats \drk in the off-road/kinodynamic environments.
It goes to show that the urban city environments~(\subref{fig:kinematic_example}) exhibit more clutter, similar to MB, and off-road environments~(\subref{fig:off_road_exp}) appear sparser as in SB.
}
\label{fig: kinematic_kinodynamic_plots}
\end{figure}
\vspace{-10pt}

\subsection{$\mathbb{R}^3,\mathbb{R}^4$ Environments}\label{subsec: R3R4}
\begin{figure}[!htb]
  \centering
  \includegraphics[width=\linewidth]{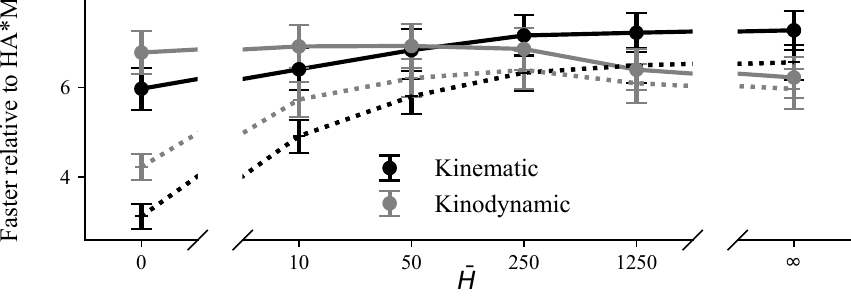}
  \caption{\textit{When} \ihastar terminates before reaching the expansion limit, \ighastar variants generally terminate faster.
  The solid lines represent the speed-up in terms of expansions, whereas the dotted lines represent the wall-clock speed-up for our C++/CUDA implementation.
  Observe that the speed up in expansions (solid lines) is not hypersensitive to the hysteresis value. 
  Expansions take $\approx50\mu s$ on a desktop~(Intel i9-12900k CPU/Nvidia RTX3080 GPU), whereas projecting each vertex takes $\approx0.3\mu s$.
  When $\bar{H}$ is small, switchbacks occur more frequently, leading to more time spent projecting and activating vertices.
}
  \label{fig:H3_figure}
  \vspace{-15pt}
\end{figure}

Here, we intend to show that the best path performance of \ighastar variants is not hypersensitive to the choice of the hysteresis threshold hyperparameter.
We consider two types of environments: urban city maps (from the moving AI benchmark~\citep{moving_AI_benchmark}, example shown in Fig.~\ref{fig:kinematic_example}) with a kinematic car~\citep{kinematic_flat} ($\mathbb{R}^3$), 
and off-road maps (from BeamNG~\citep{beamng_tech}, example shown in Fig.\ref{fig:off_road_exp}) with a kinodynamic car~\citep{non_planar_analysis}, assuming no side-slip, with its state~($\mathbb{R}^4$) defined by its $\mathrm{SE}(2)$ pose and body frame forward velocity.
The other components of the state (roll, pitch, and so on) are unique for a given $\mathrm{SE}(2)$ pose, velocity, and elevation map.
We generate 500 queries for both environments with the hyperparameters (related to successor, goal set, starting resolution) set such that \hastar can find \textit{a} solution at the starting resolution to reflect a practical setting.
We use variants of \ighastar with the $\bar{H}$ drawn from $[0, 10, 50, 250, 1250, \infty]$
and compare them to \ihastar.
We use $\bar{H}=0,\infty$ as references, and chose the intermediate values for $\bar{H}$ based on the average expansions used by \ihastar to find the first solution for both environments, which is $\sim1500$.
All methods are limited to 100,000 expansions for both environments.
Note that here we are running the planner \textit{open loop}.

\textbf{Hypothesis H3:} \ighastar-n variants use fewer expansions to terminate on average compared to \ihastar, when \ihastar terminates before the expansion limit.

\textbf{Result:} Fig.~\ref{fig:H3_figure} \textbf{confirms H3} for both the kinematic and kinodynamic environments, showing that in general the \ighastar-n variants are $\sim6\times$ faster than \ihastar.
See Table~\ref{tab:H1_H2_table} and the caption of Fig.~\ref{fig: kinematic_kinodynamic_plots} for observations on the structure in the kinematic and kinodynamic off-road planning problems.

\begin{figure*}[!htb]
\centering
\begin{subfigure}{.18\linewidth}
  \includegraphics[width=\linewidth]{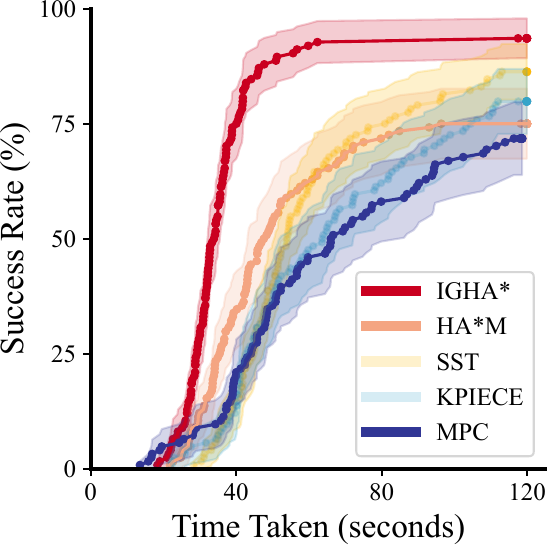}
  \caption{Simulation results}
  \label{fig:H4_figure}
\end{subfigure}%
\begin{subfigure}{0.27\linewidth}
  \includegraphics[width=\linewidth]{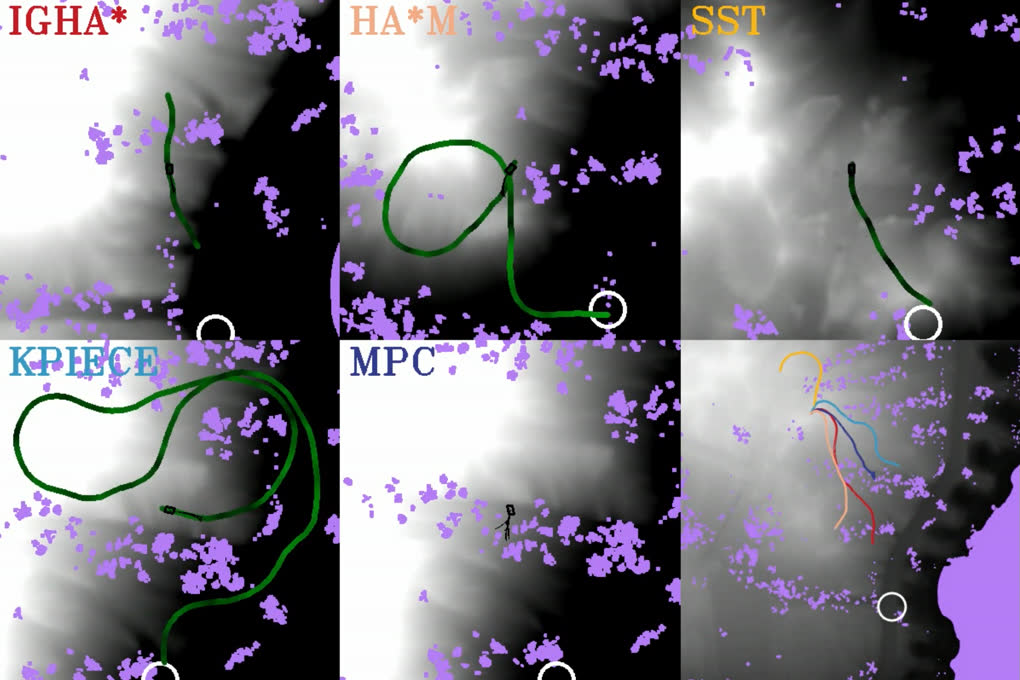}
  \caption{Anecdotal sim. example}
  \label{fig:H4_anecdote}
\end{subfigure}
\begin{subfigure}{0.18\linewidth}
  \includegraphics[width=\linewidth]{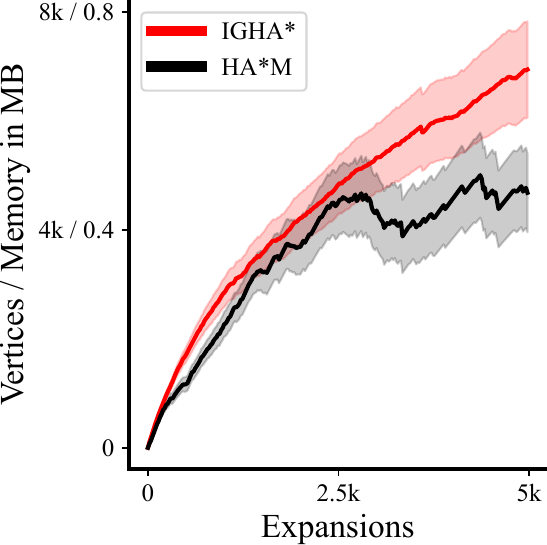}
  \caption{Memory comparison}
  \label{fig:mem_comp}
\end{subfigure}
\begin{subfigure}{.345\linewidth}
  \includegraphics[width=\linewidth]{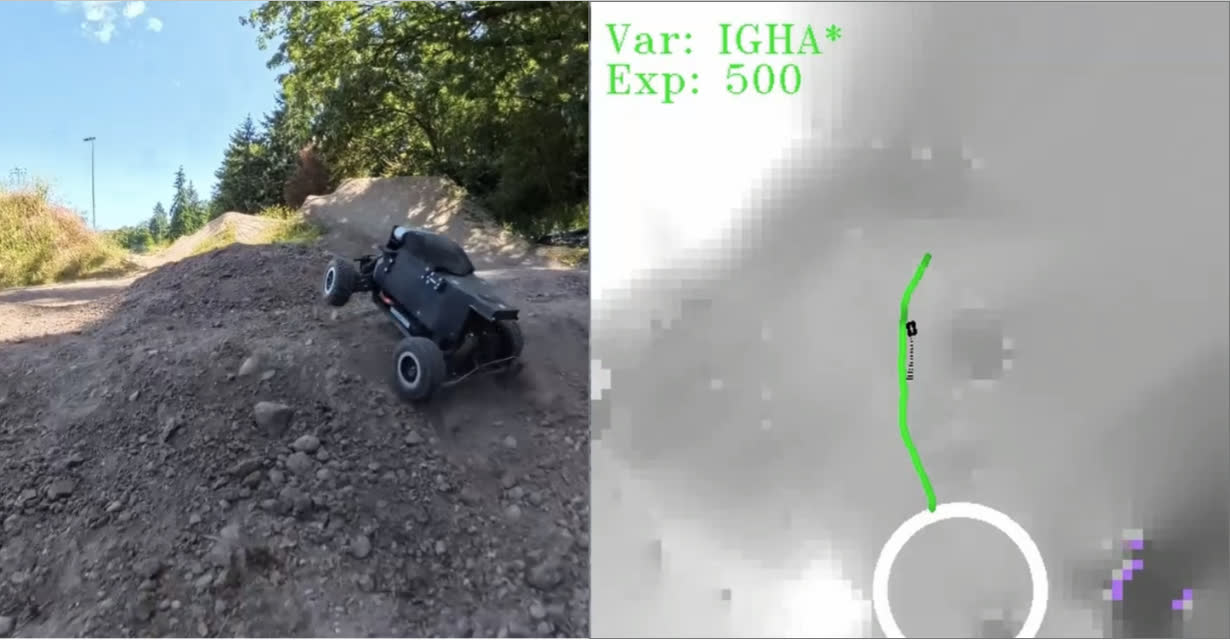}
  \caption{Small-scale robot deployment}
  \label{fig:real_world_ros}
\end{subfigure}
\caption{
Across off-road scenarios (Fig.~\ref{fig:off_road_exp}), \ighastar enables faster goal-reaching, 
outperforming \ihastar, MPC, SST, and KPIECE (\subref{fig:H4_figure}).
Representative example shown in ~\subref{fig:H4_anecdote}, where the heightmap is grayscale (lighter = higher), with purple as obstacles, green as the plan (brighter = faster), black for MPC rollouts, and a white circle marking $V_g$.
\ighastar (top left) yields lower-cost plans than \ihastar, SST, and KPIECE, (top center, top right, bottom left resp.).
Both \ighastar and \ihastar outperform MPC (bottom center) in avoiding local minima (global snapshot shown bottom right).
Here, \ighastar uses $\approx0.8$MB (~\subref{fig:mem_comp}, 200 samples) for the setup corresponding to \subref{fig:H4_figure}, 
well within the capacity of embedded systems such as the Jetson Orin NX 16GB.
Real-world tests~(\subref{fig:real_world_ros}) in an outdoor dirt bike park demonstrate online planning on a small-scale platform(left) at 3-4 Hz under tight computational limits, using a $16\times16m^2$ body-centric map (right).
}
\label{fig:In_the_loop_exp}
\vspace{-15pt}
\end{figure*}

\subsection{Closed Loop Experiments for Off-Road Autonomy}\label{subsec: closed_loop_exp}

We run the kinodynamic planner in the loop with MPPI~\citep{williams2017model}, using a real-time C++/CUDA implementation with GPU-accelerated edge expansion and evaluation.
Fig.~\ref{fig:H3_figure} shows that the wall-clock speedup (dotted grey lines) peaks near the middle, which aids in tuning for the hysteresis value.
To avoid perception and state estimation confounds, we run experiments in BeamNG~\citep{beamng_tech}, accessing ground-truth state and body-centric maps over a $100{\times}100\,\text{m}^2$ area at 25 Hz.
We compare \ighastar-100 (or \ighastar), \ihastar, SST~\citep{SST} and KPIECE~\citep{kpiece} from OMPL~\citep{ompl}, and a baseline MPC.
The MPC uses a 1.5s horizon with 1024 rollouts when used with the planner, but 5s with 8192 rollouts when run as a baseline for it to stay competitive.
We tune the resolution for SST to avoid overpruning at standstill, using the same resolution for KPIECE for consistency, and default values for other hyperparameters.
Hyperparameters for map size, MPC horizon, control frequency, and so on follow off-road autonomy norms~\citep{han2023model, meng2023}, with a planning budget of 5000 expansions, replanning at $\approx$4~Hz.  
We evaluate 30 start-goal pairs ($\approx$250m apart), each run 4 times with a 120s timeout; success requires reaching the goal within time.  
If the goal lies outside the map, it's projected to the edge.

\textbf{Hypothesis H4:} \ighastar has a higher success rate compared to the baselines.

\textbf{Result:}
Fig.~\ref{fig:H4_figure} shows that \ighastar generally achieves higher success rates than the baselines, and reaches the same success rates within a much shorter time frame.
Fig.~\ref{fig:H4_anecdote} illustrates this: \ighastar yields lower-cost paths than the baselines under a fixed budget.

\textbf{Deployment:}
The kinodynamic planner runs at $\approx$20k, 4k, and 2k expansions/sec on our desktop, Nvidia Orin AGX (32GB), and Orin NX (16GB), respectively, yielding a $\approx$4 Hz planning rate in simulation on desktop,
(vertex-mem. req. shown in Fig~\ref{fig:mem_comp}).
We deploy on a small-scale off-road autonomy platform with all modules (MPC, perception, planning) running onboard on an Orin NX at $\approx$2 m/s in real off-road terrain (Fig.~\ref{fig:real_world_ros}; see~\citep{talia2023demonstrating} for details about the platform, MPC, and the perception system).  
On-robot, we target a 3–4 Hz update rate with a budget of 500 expansions
(actual:$4.4\pm0.3$Hz, variations due to intermittent resource contention and early termination by the planner).

%% file: sections/Future_Work.tex
\section{Discussion}
\label{sec:discuss}

\ighastar represents a shift from a purely expansion-driven planner to a more general \emph{vertex reactivation framework}, capable of leveraging both prior computation and resolution-aware reasoning.
As long as the implementation of alternative \shift and \activate methods fit within the definitions in Sec.~\ref{sec:theory}, the guarantees hold.
Notice that \ihastar can be constructed using \ighastar, by using a \shift method that always increases resolution but never breaks the loop, and an \activate method that resets $Q_v$ with $v_s$.
While a large number of variations of \ighastar are possible, the focus of this work is to show the value that even a straightforward variant can bring.
Further, note that in our deployment, we make no domain-specific modifications to Alg.~\ref{alg:igha}; deploying the algorithm on a different robot, or for higher-dimensional environments, is a matter of choosing the appropriate successor, heuristic, and goal set definitions -- computational burden, due to the curse of dimensionality in higher dimensional problems, is expected to change across the board.

\textbf{Remark:} in the instantiation in Sec.~\ref{subsec: ighastar-H}, the rule used is \textit{naive}; it only depends on information directly available in $Q_v$; the $\bar{H}$ we use for our closed loop experiments~(Sec.~\ref{subsec: closed_loop_exp}) is informed by our domain knowledge about off-road autonomy.
However, for different domains, one could learn the appropriate \textit{average} $\bar{H}$ values or predict them on the fly, while technically still retaining the guarantees in Sec.~\ref{sec:theory}.
While we use a grid-based discretization approach to perform duplicate detection, other methods of discretization, such as use of r-disks as in~\citep{SST, Sparse_RRT} or successor validity as in~\citep{soft_duplicate, RCS_needle} for duplicate detection may also be used as long as they satisfy the assumptions in Sec.~\ref{sec:theory} and expose a hyperparameter equivalent to the discretization resolution used here.

\textbf{Implementation}: The pseudocode shown here omits certain mechanisms for simplicity.
We only maintain $Q_v.\text{Active}$ as a queue, and the rest as an unordered list so as to not incur the cost of sorting inactive vertices at every step of forward search.
We handle the edge cases that prevent \shift from being called at least once each iteration.
The vertex attributes, such as $g,~f$ values, and the lowest dominant resolution are set implicitly during vertex expansion.